\PassOptionsToPackage{table}{xcolor}
\documentclass{article}

\usepackage{iclr2027_conference,times}

\usepackage{amsmath,amsfonts,bm}

\def\eqref#1{equation~\ref{#1}}

\def\1{\bm{1}}

\DeclareMathAlphabet{\mathsfit}{\encodingdefault}{\sfdefault}{m}{sl}
\SetMathAlphabet{\mathsfit}{bold}{\encodingdefault}{\sfdefault}{bx}{n}

\usepackage{hyperref}
\usepackage{url}
\usepackage{xcolor}
\usepackage{enumitem}
\usepackage{booktabs}
\usepackage{graphicx}
\usepackage{caption}
\usepackage{subcaption}
\usepackage{wrapfig}
\usepackage{float}
\usepackage{algorithm}
\usepackage{algorithmic}
\usepackage[normalem]{ulem}
\usepackage[percent]{overpic}
\usepackage{tikz}
\usepackage{comment}\usepackage{fontawesome5}
\definecolor{coreblue}{HTML}{D9EAF3}

\title{Procedural Core: A Compact Recurrent\\Initialization for Vision Transformers}

\author{%
\textbf{Zachary Shinnick}$^{1,*}$ \qquad
\textbf{Christian Intern\`o}$^{2}$ \qquad
\textbf{Hemanth Saratchandran}$^{1}$\\[0.15em]
\textbf{Anton van den Hengel}$^{1,3}$ \qquad
\textbf{Damien Teney}$^{4}$\\[0.6em]
{\normalfont\normalsize
$^{1}$Australian Institute for Machine Learning (AIML), Adelaide University}\\[0.15em]
{\normalfont\normalsize
$^{2}$Bielefeld University \qquad
$^{3}$Metacognition AI \qquad
$^{4}$Idiap Research Institute}\\[0.4em]
{\normalfont\small
$^{*}$Correspondence: \texttt{zachary.shinnick@adelaide.edu.au}}%
}

\iclrfinalcopy 
\begin{document}

\maketitle
\begin{abstract}
Transformers are typically trained from random initialization, requiring all their capabilities to emerge from large-scale optimization. Recent work showed that a small amount of abstract procedurally generated
data can help acquire
generic inductive structure at low 
cost. However, this adds a pretraining stage that must be repeated for every target model.
We propose \textit{Procedural Core}, an initialization strategy that captures this generic structure into a compact set of weights that can be reused across models.
We train a minimal recurrent transformer on 
procedural data, then 
expand its weights to initialize transformers of arbitrary width and depth. The resulting initialization improves performance on image classification, self-supervised visual learning (DINO), and
modeling natural language (\textsc{FineWeb-Edu}) and code (\textsc{CodeParrot}).
For image classification, expanding a 1M-parameter core to initialize an 85M-parameter ViT-Base improves
\textsc{ImageNet}
top-1 accuracy by 2.2\,pp
over standard random initialization. Our analysis identifies recurrence as essential for learning compact weights that transfer across models.
In ViTs,
we localize a key benefit in the suppression of high-norm tokens
that produces substantial improvements in zero-shot segmentation (\textsc{\mbox{ImageNet-S}} mAP 32.3\,→\,42.9), object localization (\textsc{VOC07} CorLoc 9.9\,→\,18.4), and depth estimation (\textsc{NYUv2} RMSE 1.104\,→\,0.998).
This demonstrates that transformers need not start from a blank slate,
and can be initialized with generic capabilities at low cost with no domain- or task-specific data.
\par\smallskip
\noindent
\faGlobe\enspace\textbf{Project page:}\enspace
{\footnotesize
\href{https://zlshinnick.github.io/procedural-core/}{%
  \nolinkurl{zlshinnick.github.io/procedural-core/}}
}
\end{abstract}

\begin{figure}[h!]
    \centering
    \includegraphics[width=0.99\linewidth]{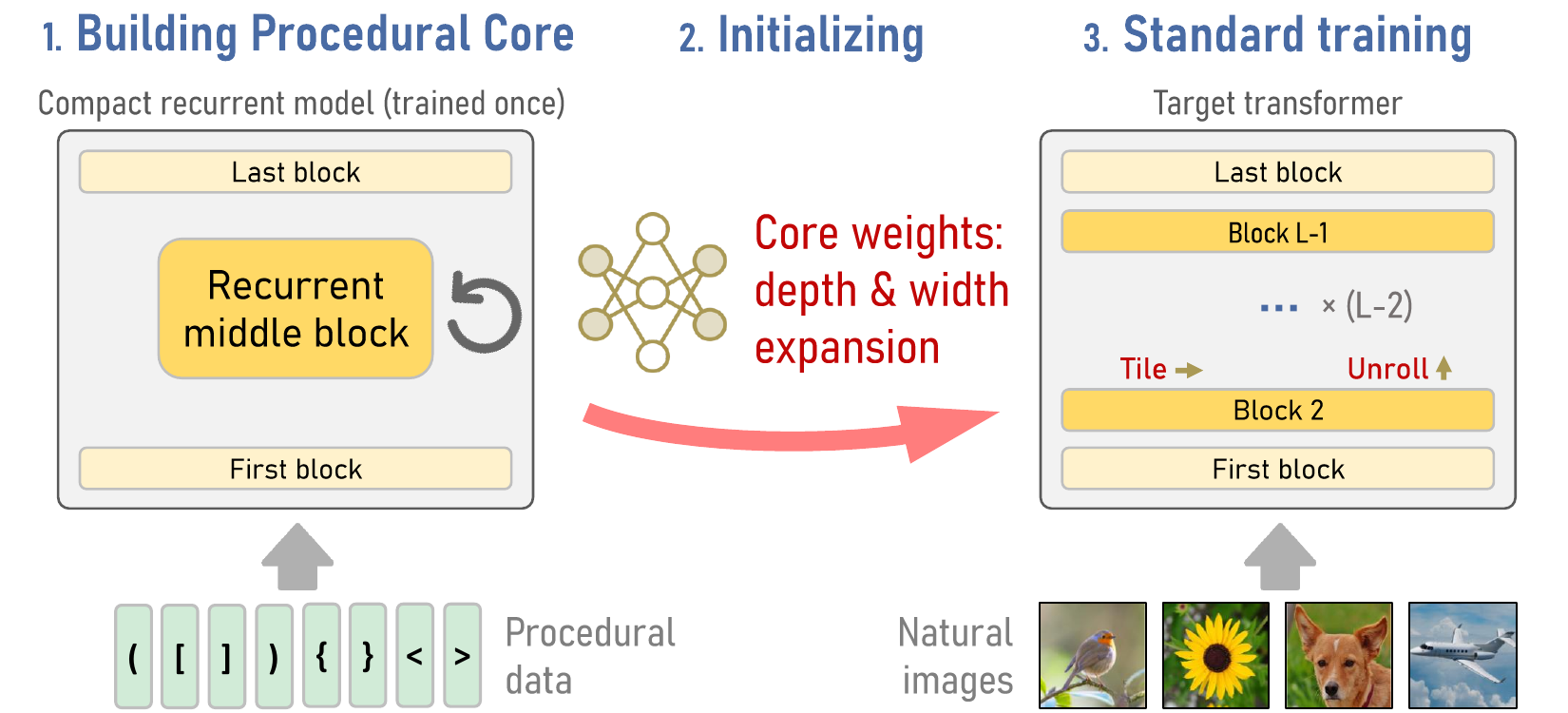}
    \vspace{1pt}
    \caption{
    We propose a generic initialization for transformers
    as an alternative to random weights.
    \textbf{(Left)~We first train a small auxiliary model} to capture generic computations. We generate abstract training data with simple algorithms
    such as formal languages with nested structures
    (pictured as balanced parentheses).
    The model is recurrent in depth
    such that it captures regularities of the data
    in a minimal set of weights: the \textit{Procedural Core}.
    \textbf{(Middle)~We unroll and expand the Procedural Core}
    to initialize a target model
    of arbitrary width and depth.
    \textbf{(Right)} This bootstraps the subsequent standard training,
    improving convergence and generalization compared to a random initialization.
    }
    \label{fig:teaser}
    \vspace{-16pt}
\end{figure}
\section{Introduction}

Recent work on large pretrained transformers shows
that they learn
mechanisms \citep{han2025learning,lu2022frozen}
and representations \citep{buzeta2026seeing,huh2024position,gupta2025better,riachi2025random} 
that transfer across tasks and modalities.
This suggests that they contain some internal structure that is generic and broadly reusable.
Motivated by this perspective, a growing body of work explores pretraining on
abstract data generated procedurally,
as an efficient way to acquire generic structure in the model weights \citep{zhang2024intelligence, hu2025between, jiang2026proceduralpretraining, shinnick2025transformers, shinnick2025learnwithoutimages}.
Such a ``\textit{procedural~warm-up}'' uses data that is free of
semantic or visual content, and this strategy can be strikingly data-efficient.
In ViTs,
\citet{shinnick2025learnwithoutimages}
show that allocating
1\% of the training budget initially to non-visual procedural data%
\footnote{\textit{Procedural data} refers to data produced by
simple algorithmic rules such as formal grammars.}
can yield performance gains comparable to increasing natural image data by 28\%.
In language models, \citet{jiang2026proceduralpretraining}
show that allocating 0.1\% of the training budget initially to
procedural data enables models to achieve the same loss while consuming just 55\% of the natural language data.

Unfortunately, existing strategies with procedural data
are not a simple drop-in replacement for standard random initialization.
They require a dedicated training stage that needs to be repeated for each target model.
The effects of this stage are tightly coupled with a specific architecture and model size.
Note however that procedural data is, by definition, generated by simple algorithms
and thus has a small description length (i.e.\ low Kolmogorov complexity).
We therefore posit that the structure induced in the weights should also admit a compact representation.
In principle, it may thus be possible to obtain the same effects
with a simple deterministic procedure applicable to any target model.
The central question of this work is therefore as follows: \textbf{\emph{can the useful structure acquired by training on procedural data
be compressed into a compact, transferable core?}}

To address this question, we hypothesize that 
the mechanisms in models trained on procedural data
could be implemented more compactly through \textit{recurrence}.
Existing models like Universal Transformers~\citep{dehghani2019universaltransformers} already show that depth-wise recurrence can capture expressive computations
with a small set of parameters, and recent work also shows that recurrence can benefit reasoning tasks~\citep{saunshi2024inductivebiasstackingimproving}. The Block-Recurrent Hypothesis~\citep{jacobs2025block} further suggests that deep transformers, including ViTs, approximate compact recurrent programs.
This implies that multi-layer computations could often be represented more compactly with few reusable blocks.
Together, these works indicate that recurrence is a natural mechanism for consolidating
the structure in trained transformers into a compact representation.

Building on this perspective, we propose \emph{Procedural Core}: a compact set of weights obtained by training a minimal recurrent auxiliary model
on procedural data. More precisely, we train a small recurrent ViT with three unique blocks (input, recurrent middle block, output) and then use its weights to initialize larger architectures by unrolling the recurrent block in depth and tiling it in width (see Figure~\ref{fig:teaser}). This parameter sharing across layers forces the auxiliary model to encode
its computations into a compact set of weights, acting as a regularizer and
facilitating their reuse.
Concretely, we now have a generic strategy to initialize useful structure in the weights of a transformer of any size, in a single deterministic step.
Our choice of procedural data and training objective follows prior work~\citep{shinnick2025learnwithoutimages,jiang2026proceduralpretraining}.
Our primary objective here is therefore to obtain comparable effects with a much more practical procedure.

We evaluate \textit{Procedural Core} across model sizes and domains.
We get consistent gains over random and structured initializations~\citep{trockman2023mimetic}.
For example, we expand our $\sim$1M-parameter core to initialize an 85M-parameter ViT-Base,
then train it on \textsc{ImageNet-1K}~\citep{deng2009imagenet}. This yields a +2.2 percentage point improvement in top-1 accuracy compared to a random initialization.
Importantly, these gains extend to other applications, including self-supervised vision models (DINO) and language models trained on natural language (\textsc{FineWeb-Edu} dataset,~\citealt{penedo2024the}) and code (\textsc{CodeParrot} dataset,~\citealt{codeparrot}).
These results suggest that the method instantiates generic structure
in the weights that is broadly useful across domains.

We analyze why the \emph{Procedural Core} transfers.
We find that depth-wise parameter sharing distributes computation across
a wider set of directions, including lower-energy components important
for downstream transfer.
In ViTs, this structure concentrates in the attention
value/output pathway, where it suppresses high-norm token outliers and improves
representations across segmentation, localization, and depth estimation.
Our contributions are summarized as follows.

\textbf{(1) An initialization strategy} (\emph{Procedural Core})
that instantiates useful structure in the weights of a transformer of any size, as an alternative to standard random weights.
It uses a compact set of weights obtained by training
a small recurrent model on procedural data (Section~\ref{sec:4}).

\textbf{(2) An empirical evaluation} of the proposed strategy for computer vision and language.
On \textsc{ImageNet-1K}, Procedural Core improves ViT-Base top-1 accuracy by +2.2 percentage points over standard random initialization.
The benefits extend beyond classification,
with improvements across a  wide array of tasks including
semantic segmentation
(\textsc{ADE20K}: 26.6$\,\rightarrow\,$28.8 mIoU), zero-shot segmentation
(\textsc{ImageNet-S}: 32.3$\,\rightarrow\,$42.9 mAP), object localization
(\textsc{VOC07}: 9.9$\,\rightarrow\,$18.4 CorLoc), and depth estimation
(\textsc{NYUv2}: 1.104$\,\rightarrow\,$0.998 RMSE).
We also observe gains with self-supervised DINO models and language models trained on natural language and code
(Section~\ref{sec:5}).

\textbf{(3) A mechanistic analysis} of the benefits.
We find that the recurrence distributes useful computations across a larger set of dominant directions.
We also identify the attention value/output pathway as carrying much of the transferable structure, which we connect to the suppression of high-norm tokens and improvements in segmentation, localization, and depth estimation
(Section~\ref{sec:analysis}).


\section{Related Work}

\textbf{Initialization of vision transformers.}
ViTs are typically initialized with random weights~\citep{glorot2010understandingthedif,he2016deepresiduallearning}. Several works propose 
handcrafted patterns in attention layers e.g.\ to instill a convolution-like prior
\citep{huang2020improving,zhao2022zero,trockman2023mimetic,zheng2025structured,giri2025ibitutilizinginductivebiases}.
Other approaches transfer structure from pretrained models,
by distilling shared weight templates
or expanding the weights of smaller models~\citep{feng2025wave,xu2023initializing,chen2022bert2bert,samragh2024scaling,panigrahi2024efficient}.
In comparison, our approach transfers structure from an auxiliary model
that provides compact, transferable weights by construction,
through its recurrent architecture and its training on abstract data.

\vspace{3pt}
\textbf{Pretraining on procedural data.}
A series of works on language modeling use data generated
with formal languages or cellular automata to learn structure independently from semantics~\citep{chiang2022transferability,mccoy2023modeling,papadimitriou2023injecting,lindemann2024sip,wu2022insights,zhang2024intelligence}.
Recent work uses such procedural data to pretrain large models
before exposure to natural language%
~\citep{hu2025between,jiang2026proceduralpretraining,shinnick2025transformers}
or images~\citep{shinnick2025learnwithoutimages}
to improve data efficiency and generalization.
While these approaches require a new initial training phase, 
we amortize this phase by summarizing its outcome in a compact set of weights that can be reused in a single deterministic step
to initialize a transformer of any size.

\vspace{3pt}
\textbf{Training vision models on abstract data.}
There is a long history 
with synthetic images of
fractals~\citep{nakamura2023pretrainingvisiontransformerslimited, nakamura2024scaling},
contours~\citep{kataoka2022replacinglabeledrealimagedatasets}, 
and structured noise~\citep{baradad2022learninglookingnoise}.
Non-visual data like text, code, and mathematics~\citep{du2025large, zheng2024lm4lv, han2025learning, huh2024position}
can also foster capabilities for visual reasoning.
Our work adds evidence for shared basic computations that transfer across tasks and modalities.

\vspace{3pt}
\textbf{Recurrence and parameter sharing in transformers.}
Parameter sharing and depth-wise recurrence have been explored to improve efficiency and computational expressivity. For example, ALBERT~\citep{lan2020albert} shares parameters across layers to reduce memory usage, while the Universal Transformer~\citep{dehghani2019universaltransformers} uses depth recurrence to enable iterative computation. Recent work further suggests that recurrence induces inductive biases beneficial for reasoning~\citep{saunshi2024inductivebiasstackingimproving}, and the Block-Recurrent Hypothesis~\citep{jacobs2025block} argues that deep ViTs approximate compact recurrent programs composed of reusable blocks.
Our method uses depth recurrence as a way to obtain a compact set of weights from the procedural pretraining, which can be expanded to an arbitrary depth in the target model.
\clearpage

\section{Proposed method}
\vspace{-1pt}
\label{sec:4}

We build on prior work that describes a \textit{procedural warm-up} for ViTs~\citep{shinnick2025learnwithoutimages}
that exposes the model to non-visual data
before standard pretraining on 
ImageNet. 
The data is generated with simple algorithms
that produce sequences of abstract tokens.
They teach the model to process 
compositional and nested patterns
prior to the semantic information of image datasets.
We adopt the data generation and training approach from \citet{shinnick2025learnwithoutimages} as summarized below.

\vspace{-1pt}
\textbf{Generating procedural data.}
We generate sequences with nested compositional patterns, which prior work has shown to be effective for training ViTs~\citep{shinnick2025learnwithoutimages}.
These sequences encode random \textit{push} and \textit{pop} operations on a stack, a last-in, first-out memory structure.
The tokens have no meaning. For visualization, 
we represent them with various brackets and parentheses where \textit{push} and \textit{pop} symbols correspond respectively to opening and closing symbols.
The sequences then resemble sequences of matching parentheses, such as:
\colorbox{gray!25}{%
  \texttt{\small(\,[\,]\,)\,\raisebox{.5pt}{\scriptsize\{\,\}}}%
}
a.k.a.\ a Dyck language~\citep{hu2025between, papadimitriou2023injecting}. 
We use a vocabulary of 128 tokens (64 matched pairs)
and sequences of length $N\!=\!H\!\times\!W$ where $H\!\times\!W$ is the size of the target ViT's input grid.

\vspace{-1pt}
\textbf{Training ViTs on non-visual data.}
We make it possible for ViTs to ingest abstract tokens
by replacing their patch embedding layer (typically a linear projection) with
an embedding layer as found in language models (i.e.\ a look-up table).
The embeddings are initialized randomly and held frozen, such that
all the learning with the procedural data happens in attention and MLP layers.
 After training on procedural data, these token embeddings are discarded, and the visual patch embeddings are then learned exclusively on standard visual data.
The training objective with procedural data is a
masked-token prediction objective.
We only mask tokens that admit a unique valid completion (i.e.\ a subset of the closing tokens).
To solve the task, the model must discover and simulate the stack-based process
to predict valid completions.
See Appendix~\ref{sec:pwu} for details.

\label{methodStepOne}

\vspace{-1pt}
\textbf{Overview of our method.}~
The \textit{procedural warm-up}
method of
~\citet{shinnick2025learnwithoutimages} required
an extra training phase to be repeated
for every target architecture and size.
Instead, we train a small auxiliary model on procedural data (Section~\ref{sec:learning_core_weights}) and expand its learned weights to initialize a target model of arbitrary depth and width (Section~\ref{sec:4.2}). As we show in
Section~\ref{sec:5},
this approach promotes the learning of transferable structure. It also makes procedural initialization more accessible. Once learned, the core weights can initialize new models in a single efficient deterministic step, eliminating the need to repeat the procedural training.

\vspace{-5pt}
\subsection{Learning the Core Weights}
\vspace{-1pt}
\label{sec:learning_core_weights}

We construct our auxiliary model as a ViT-Tiny with the 
embedding modification described in Section~\ref{sec:4}. To introduce recurrence, we tie all weights of blocks $2$~to~$L$\,-\,$1$, where $L\!=\!12$ is the number of blocks (one block refers to a pair of attention and MLP layers).
The first and last blocks remain untied, as prior work shows that they perform specialized computations~\citep{mcleish2025teaching,skean2025layer}. Our initial experiments also showed degraded performance with all layers tied. 
The resulting recurrent model is trained on procedural data following
\citet{shinnick2025learnwithoutimages}.

The recurrence serves first
as a regularizer, preventing layer-wise specialization.
This encourages the model to learn general, transferable computations. Second, the recurrence produces a compact set of parameters that can be replicated to an arbitrary depth model, as described in Section~\ref{sec:4.2}.

\vspace{-4pt}
\subsection{Expanding the Core Weights to Arbitrary Depth and Width}
\vspace{-1pt}
\label{sec:4.2}

After training the auxiliary model,
we transfer
the weights of its first, recurrent middle, and last blocks
to initialize a target model 
of arbitrary (typically larger) depth and width.

\vspace{-2pt}
\setlength{\abovedisplayskip}{1pt}
\setlength{\belowdisplayskip}{3pt}
\textbf{Depth expansion.}
To initialize a model of depth $\tilde{L}$, we copy the first and last
blocks and replicate the recurrent middle block
$\tilde{L}\!-\!2$ times.
The middle blocks are initialized
identically but they are untied and free to diverge during the subsequent training. 

\textbf{Width expansion.}
To initialize a wider model, we tile each matrix of core weights
$W\in\mathbb{R}^{d_{\rm out}\times d_{\rm in}}$ as follows,
rescaling the result to preserve
its Frobenius norm:
\begin{equation}
\widetilde W_{ij}
= W_{(i\bmod d_{\rm out}),(j\bmod d_{\rm in})},
\qquad
\widetilde W ~~\leftarrow~~
\widetilde W ~\, \|W\|_F \,\big/\, \|\widetilde W\|_F.
\end{equation}
Concatenated attention weights
are split into their constituent
$Q$, $K$, and $V$ matrices before expansion. Non-matrix parameters use
neutral padding: $0$ for biases and $1$ for normalization scales.

\section{Experiments}
\label{sec:5}
\setlength{\tabcolsep}{6pt}


\vspace{-4pt}
\subsection{Supervised Vision Transformers}
\vspace{-1pt}
\label{sec:5.1}

\textbf{Motivation.}
Prior work has demonstrated that
an initial training phase on procedural data (or \textit{warm-up})
helps a ViT-B converge faster
in a subsequent 
standard supervised training on ImageNet~\citep{shinnick2025learnwithoutimages}.
We now evaluate whether our initialization
provides similar benefits.

\textbf{Experimental setup.}
We initialize a ViT-Base (85M parameters) with our method (expanding a set of $\sim$1M core weights) and train it for 300 epochs on \textsc{ImageNet-1k}
following a standard protocol and hyperparameters~(\citealt{xu2023initializing}; see Appendix~\ref{sec:abs_training_detail_supervised} for details). 
We compare our initialization with three alternatives:
the \textit{default initialization} of ViTs with random weights from a truncated-normal distribution~\citep{dosovitskiy2020image};
the \textit{Mimetic initialization}
that imitates the diagonal attention patterns of trained ViTs
with a handcrafted 
formula~\citep{trockman2023mimetic};
and  the \textit{Procedural warm-up}
that directly pretrains the target model on the same procedural data
as used with our method~\citep{shinnick2025learnwithoutimages}.

\begin{figure}[h]
    \centering
    \begin{minipage}[c]{0.47\linewidth}
        \renewcommand{\arraystretch}{1.17}
        \centering
        {\small
        \setlength{\tabcolsep}{4pt}
        \begin{tabular}{lcc}
        \toprule
        & ~Top-1 \small(\%) & ~~~~$\Delta$ \small(\%) \\
        \midrule
        Default initialization
        & ~~77.6 \scriptsize$\pm$ 0.2 & +0.0 \\
        
        Mimetic initialization
        & ~~79.5 \scriptsize$\pm$ 0.7 & +1.9 \\
        
        Procedural warm-up
        & ~~79.4 \scriptsize$\pm$ 0.3 & +1.8 \\

        \cellcolor{coreblue}\textbf{Procedural Core}
        \rule{0pt}{10pt} 
        & ~~\cellcolor{coreblue}\textbf{79.8} \scriptsize$\pm$ 0.6
        & \cellcolor{coreblue}\textbf{+2.2} \\[-1.6pt]
        \bottomrule
        \end{tabular}
        }
        \vspace{19pt}
    \end{minipage}
    \hfill
    \begin{minipage}[c]{0.50\linewidth}
        \centering
        \begin{tikzpicture}
          \node[anchor=south west, inner sep=0] (img) at (0,0) {%
            \includegraphics[
              width=\linewidth,
              trim=0 20pt 0 0,
              clip
            ]{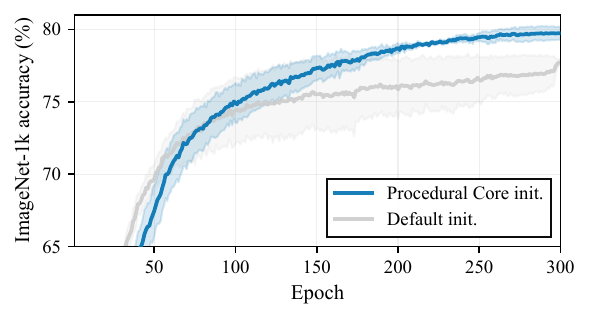}%
          };
          \node[anchor=south west, xshift=17pt, yshift=-2.9pt] at (img.south west)
            {\scriptsize Epoch\,$\rightarrow$};
        \end{tikzpicture}
    \end{minipage}
    \vspace{-9pt}
    \caption{\textbf{Procedural Core improves ViT-Base training on
    \textsc{ImageNet-1k}.}
    \textbf{Left:} Final top-1 accuracy (mean $\pm$ s.d. over three seeds).
    \textbf{Right:} Mean training trajectories over the same seeds, showing a
    persistent advantage over default initialization.}
    \label{fig:imagenet_results}
\end{figure}

\textbf{Results.}
Figure~\ref{fig:imagenet_results} 
shows that our initialization improves
the top-1 accuracy over the default initialization by +2.2\% and outperforms the other alternatives.
It surpasses in particular the procedural warm-up both here and on
\textsc{CIFAR-100} (Appendix~\ref{app:warmup-comparison}).
This supports the suitability of our small recurrent 
model
to capture reusable structure from the procedural data.
Figure~\ref{fig:imagenet_results} (right) also shows that the
advantage persists throughout training rather than providing only an initial
head start.

\vspace{-7pt}

\vspace{-3pt}
\textbf{Downstream tasks.}
We evaluate in Appendix~\ref{app:cifar-transfer} whether the gains persist beyond pretraining on \textsc{ImageNet-1k}.
The results show that the advantage persists after fine-tuning on \textsc{CIFAR-100}.
Section~\ref{sec:downstream}
will show broader 
gains for segmentation, localization, and depth estimation.

\vspace{-1pt}
\subsection{Self-Supervised Vision Transformers}
\vspace{-1pt}

\textbf{Motivation.}
Having shown in Section~\ref{sec:5.1} that Procedural Core improves large-scale supervised training, we now investigate whether these benefits extend to
self-supervised vision models.

\textbf{Experimental setup.}
We initialize a ViT-Small with our method and train it
following DINO~\citep{caron2021emerging}
for 300 epochs on \textsc{ImageNet-1k}.
We evaluate the DINO representations with $k$-NN classification below
and with linear probing in Appendix~\ref{sec:linear-probe}.
Our baseline is a standard DINO model that we trained
in the same standard conditions except for a random initialization
(see Appendix~\ref{sec:abs_training_detail_selfsupervised} for details).

\vspace{-10pt}
\begin{figure}[h]
    \centering
    \begin{minipage}[c]{0.43\linewidth}
        \renewcommand{\arraystretch}{1.17}
        \centering
        {\small
        \setlength{\tabcolsep}{4pt}
        \begin{tabular}{lccc}
        \toprule
        & \multicolumn{3}{c}{$k$-NN accuracy \small(\%) } \\
        \cmidrule{2-4}
        Epoch & 100 & 200 & 300 \\
        \midrule
        Default
        & 66.8 & 69.7 & 72.0 \\
        \cellcolor{coreblue}\textbf{Proc.\ Core}
        \rule{0pt}{10pt} 
        & \cellcolor{coreblue}\textbf{67.4} {\scriptsize(+0.6)}
        & \cellcolor{coreblue}\textbf{70.1} {\scriptsize(+0.4)}
        & \cellcolor{coreblue}\textbf{72.3} {\scriptsize(+0.3)} \\[-1.6pt]
        \bottomrule
        \end{tabular}
        }
        \vspace{13pt}
    \end{minipage}
    \hfill
    \begin{minipage}[c]{0.50\linewidth}
        \centering
        \begin{tikzpicture}
          \node[anchor=south west, inner sep=0] (img) at (0,0) {%
            \includegraphics[
              width=\linewidth,
              trim=0 20pt 0 0,
              clip
            ]{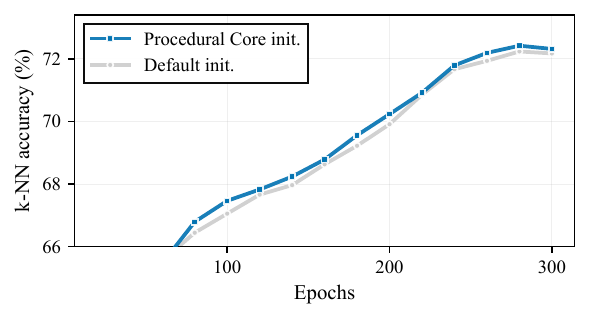}%
          };
          \node[anchor=south west, xshift=24pt, yshift=-3.1pt] at (img.south west)
            {\scriptsize Epoch $\rightarrow$};
        \end{tikzpicture}
    \end{minipage}
    \vspace{-8pt}
    \caption{\textbf{Procedural Core improves representation learning with DINO  on \textsc{ImageNet-1k}.}
    \textbf{Left:} $k$-NN accuracy at selected epochs. \textbf{Right:} $k$-NN accuracy throughout training.}
    \label{fig:dino_knn}
\end{figure}

\clearpage
\textbf{Results.}
Figure~\ref{fig:dino_knn} 
shows that our initialization slightly improves $k$-NN accuracy
by +0.6\% at epoch 100 and +0.3\% at epoch~300.
Although the advantage tends to narrow, it does persist
as seen from the training curves. 
These results indicate that our method is not limited to a supervised setting.


\subsection{Transformer-Based Language Models}

\textbf{Motivation.}
The original objective of Procedural Core
is to instill generic structure in the model weights.
The benefits should therefore extend beyond vision.
We test this hypothesis by applying our method to initialize
language models that we train on natural language 
and computer code.

\begin{wrapfigure}[12]{r}{0.378\linewidth}
    \centering
    \vspace{-16pt}
    \includegraphics[width=\linewidth]{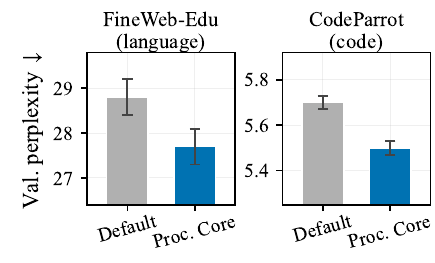}
     \vspace{-20pt}
    \caption{\textbf{Procedural Core \mbox{improves} language modeling}
    compared to a default random initialization
    (lower is better; mean $\pm$\,std.\ dev.\ over 3 seeds).}
    \label{fig:lm_multidomain}
\end{wrapfigure}
\vspace{6pt}
\textbf{Experimental setup.}
The transformers used as ViTs or language models
have architectural differences,
therefore we repeat the first step of our method
(Section~\ref{methodStepOne})
to obtain a core set of weights suitable to initialize language models.
The auxiliary model is a small GPT-2-style model~\citep{radford2019language}
with tied (recurrent) middle blocks (see Appendix~\ref{sec:lm_details} for details).
We use its weights to initialize our target model, a larger 12-block model
with 124M parameters. 
We train it for 2B tokens of either \textsc{FineWeb-Edu}~(\citealt{penedo2024the}; natural language)
or \textsc{CodeParrot}~(\citealt{codeparrot}; code).
We also train these models from a default random initialization as baselines.

\vspace{5pt}
\textbf{Results.}
Figure~\ref{fig:lm_multidomain} shows that our initialization reduces
validation perplexity by about 4\% in both domains.
Although these models are small,
the training on 2B tokens corresponds to a Chinchilla-optimal scaling of $\sim$20 tokens/parameter~\citep{hoffmann2022training},
which shows that the initialization provides
a meaningful effect after extensive language pretraining.

\vspace{2pt}
\section{Analysis of Procedural Core}
\label{sec:analysis}

We have shown that \emph{Procedural Core} consistently improves training and outperforms a direct warm-up of the target model on the same procedural data.
We now examine the impact of the architecture of the auxiliary model
(Section~\ref{secAnalysisArch}),
the spectral structure induced by recurrence and its functional importance (Section~\ref{secAnalysisSpectral}),
and how the Procedural Core alters downstream computations
(Section~\ref{sec:downstream_analysis}).
We will show that Procedural Core suppresses high-norm tokens,
which helps in particular on vision tasks with dense predictions 
(Section~\ref{sec:downstream}).

\subsection{Architectural Choices and Ablations}
\vspace{2pt}
\label{secAnalysisArch}

\begin{figure*}[t]
    \centering
    \begin{minipage}{0.95\textwidth}
        \makebox[\linewidth][l]{%
            \makebox[0.25\linewidth][c]{\small a)}%
            \makebox[0.25\linewidth][c]{\small b)}%
            \makebox[0.25\linewidth][c]{\small c)}%
            \makebox[0.25\linewidth][c]{\small d)}%
        }
        \vspace{-2pt}
        \includegraphics[width=\linewidth]{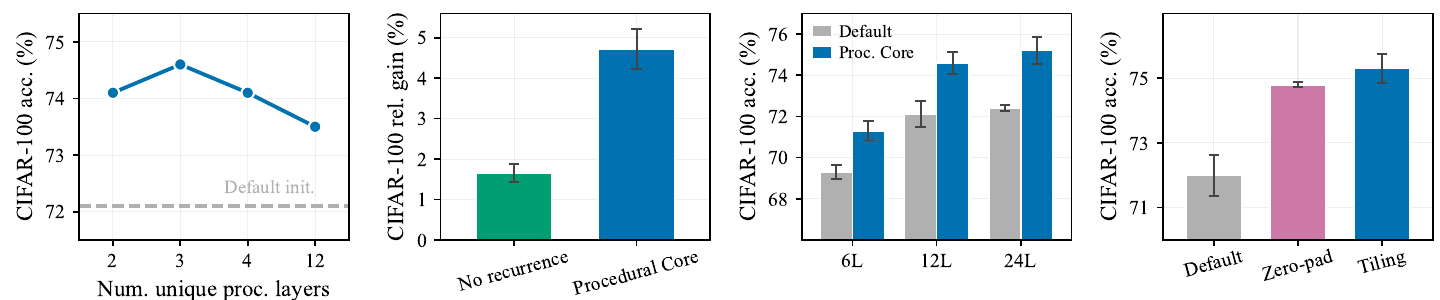}
    \end{minipage}

    \caption{\textbf{Architectural choices and ablations of our method.}
    \textbf{(a)} Transfer peaks with three unique auxiliary layers.
    \textbf{(b)} Removing recurrence substantially reduces transfer despite matching the parameter count.
    \textbf{(c)} The expansion in depth is effective with various number of layers in the target model.
    \textbf{(d)} The expansion in width is slightly more effective than a baseline with zero-padding.}
    \label{fig:core_scaling_summary}
\end{figure*}

\textbf{Recurrent parameterization.}
We assess the impact of the weight tying in the auxiliary model.
We apply our method to initialize ViT-Tiny models trained on \textsc{CIFAR-100}.
We vary the number $U$ of unique blocks in the auxiliary model while keeping the total depth fixed at $L\!=\!12$.
Figure~\ref{fig:core_scaling_summary}a shows that
transfer peaks at $U\!=\!3$, which corresponds
to the architecture used in our main experiments (independent input and output blocks plus a recurrent middle one).
Moreover, a parameter-matched 3-layer non-recurrent model yields only a $+1.7\%$ improvement over the default random initialization, compared with $+4.7\%$ for our recurrent one (Figure~\ref{fig:core_scaling_summary}b).
This highlights the importance of the compression and recurrence.

\vspace{2pt}
\textbf{Depth and width expansion.}
We evaluate the effectiveness of the proposed 
expansion
on \textsc{CIFAR-100}.
For the depth, we initialize ViT-Tiny models with varying numbers of layers as the target models.
Figure~\ref{fig:core_scaling_summary}c shows that the Procedural Core outperforms the default random initialization at every tested depth, with the gain increasing from $+2.0\%$ at 6 layers to $+2.8\%$ at 24 layers.
For the width, we initialize ViT-Small models (which are wider than our auxiliary model)
and compare the proposed tiling against a naive padding of the weight matrices with zeros.
Figure~\ref{fig:core_scaling_summary}d shows that transferring our weights with zero-padding already provides a significant gain over random weights.
The tiling yields a further benefit by filling the whole width of the network
with useful structure.
\clearpage

\vspace{-9pt}

\subsection{Spectral Structure and Its Functional Importance}
\label{secAnalysisSpectral}
\label{sec:6.1}

\textbf{The Procedural Core exhibits a broader singular spectrum.}
We now try to understand why our recurrent auxiliary model
is more effective
than a direct warm-up of the target model on the same procedural data as in \citet{shinnick2025learnwithoutimages}.
We examine weight matrices of models trained with the two methods
and compare their cumulative singular-value energy in
Figure~\ref{fig:sv-energy}.
\vspace{-2pt}


\newcommand{\legendline}[1]{%
  \textcolor[HTML]{#1}{\rule[0.35ex]{0.9em}{1.5pt}}%
}

\begin{figure}[h]
    \centering
    \includegraphics[width=0.95\linewidth, trim=0 42px 0 0, clip]{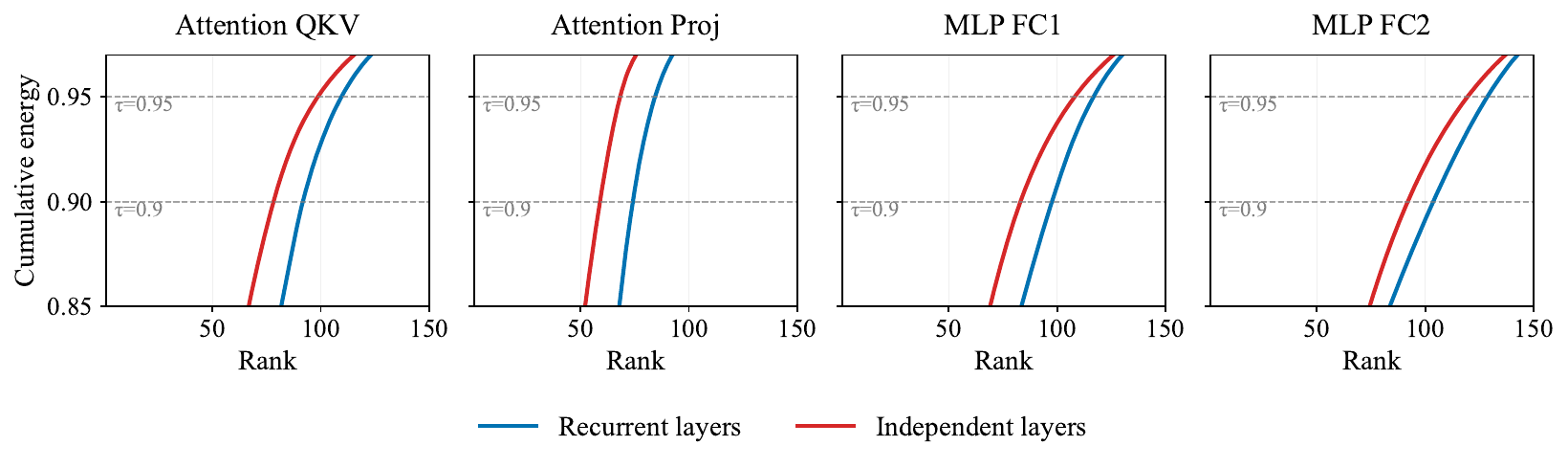}
    \vspace{-9pt}
    \caption{
    Cumulative energy of singular values of different weight matrices
    in models trained with the \textit{Procedural Core} recurrent
    architecture
    (\legendline{0675B4})
    vs.\ a direct warm-up
    (\legendline{D6292A})
    as in \citet{shinnick2025learnwithoutimages}.
    }
    \label{fig:sv-energy}
\end{figure}
\vspace{4pt}

\begin{wrapfigure}[11]{r}{0.30\columnwidth}
    \centering
    \vspace{-16pt}
    \includegraphics[width=\linewidth]{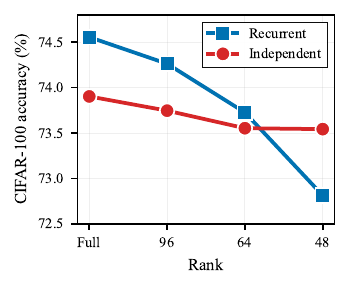}
    \vspace{-23pt}
    \caption{{Rank truncation \mbox{hinders~transfer} more sharply with Proc.\ Core weights.}}
    \label{fig:rank-trunc}
    \vspace{20pt}
\end{wrapfigure}
The \emph{Procedural Core} models
consistently show a slower spectral decay.
This can be interpreted as the computations being distributed 
across more directions in latent space.
This observation is consistent across layers (see Appendix~\ref{app:spectral} for the full results).
We also evaluate whether this property is functionally important:
we rank-truncate the weights of \emph{Procedural Core}
and \textit{warm-up} models to their top-$r$ singular directions
and use them to initialize a ViT-Tiny trained on \textsc{CIFAR-100}.
Figure~\ref{fig:rank-trunc} shows
that the warm-up weights are much less sensitive
to the truncation, whereas \emph{Procedural Core} degrades sharply
with lower ranks.
This confirms that the 
transferable computations depend on a broader set of directions,
and that \textbf{this broader spectrum is functionally important}.
\vspace{5pt}

\subsection{How Does Procedural Core Alter Computations After Training?}
\label{sec:downstream_analysis}

Having established that \emph{Procedural Core} contains functionally important structure as an initialization,
we now examine how it changes the computations in the model after training on natural images.

\begin{figure}[t]
    \centering
    \vspace{-6pt}
    \includegraphics[width=0.95\linewidth]{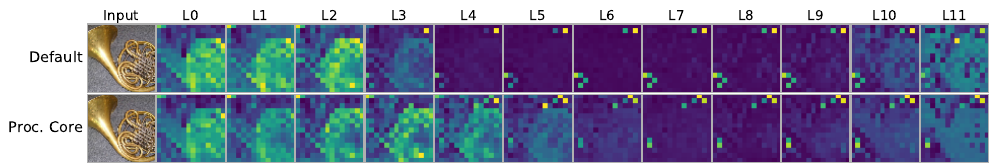}
    \vspace{-3pt}
    \caption{\textbf{Token dynamics across depth.}
    Per-token hidden-state norms for default and \emph{Procedural Core}
    initializations. High-norm outliers are suppressed from the middle blocks
    onwards with \emph{Procedural Core}. Heatmaps are independently normalized;
    extended examples are in Appendix~\ref{app:tokennorm}.}
    \label{fig:norm_maps}
    \vspace{-6pt}
\end{figure}

\textbf{Procedural Core prevents high-norm tokens.}
ViTs are known to develop outlier tokens with high norm,
often in low-information
regions~\citep{darcet2024visionregisters,jiang2026visiondontneed}.
Figure~\ref{fig:norm_maps} examines how the norms of these tokens
evolve 
in a representative image over the layers of a baseline model
(with default initialization)
vs.\ one initialized with Procedural Core.
The two models are initially similar,
but the default one develops clearer isolated high-norm outliers in middle and later
blocks, whereas \emph{Procedural Core} 
mitigates their emergence.
\clearpage

We quantify this effect on 200 images 
from \textsc{ImageNet-1k} in
Figure~\ref{fig:distributions} (details in Appendix~\ref{app:tokennorm}).
Panel~(a) tracks the bimodality of the token-norm distribution across depth,
while panels~(b,c) show the distributions at each model's peak-bimodality
block. \emph{Procedural Core} reduces peak bimodality from $1.29$ to
$0.63$ and the fraction of tokens above the high-norm cutoff from $2.20\%$
to $1.39\%$. Panel~(d) measures the attention mass received by high-norm
tokens, whose peak falls from $50.4\%$ to $17.7\%$.
Next, we identify which components drive this change.

\begin{figure}[h]
    \centering
    \vspace{7pt}
    \begin{overpic}[width=0.99\linewidth]{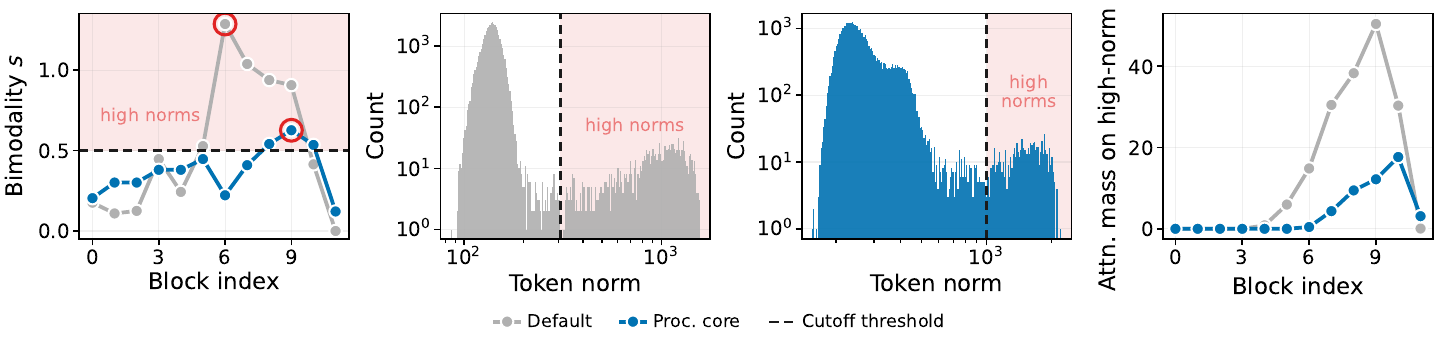}
        \put(14.5,24.7){\makebox(0,0)[c]{\scriptsize (a) Bimodality vs.\ depth}}
        \put(40.2,24.7){\makebox(0,0)[c]{\scriptsize (b) Default: $L_6$}}
        \put(65.0,24.7){\makebox(0,0)[c]{\scriptsize (c) Proc.\ Core: $L_9$}}
        \put(90,24.7){\makebox(0,0)[c]{\scriptsize (d) Attn.\ mass vs.\ depth}}
    \end{overpic}
    \vspace{-8pt}
\caption{\textbf{Procedural Core suppresses high-norm tokens and
reduces attention to them.}
Statistics on 200 \textsc{ImageNet-1k} images.
\textbf{(a)}~Bimodality across depth.
\textbf{(b-c)}~Token-norm distributions at peak-bimodality blocks.
\textbf{(d)}~Attention mass received by high-norm tokens.}
    \label{fig:distributions}
\end{figure}
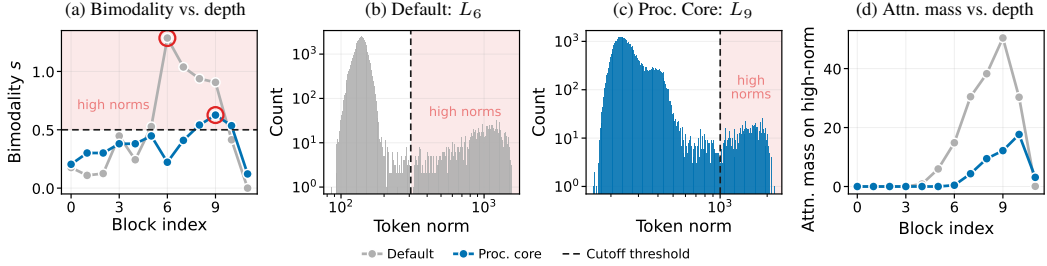
\vspace{-2pt}

\begin{figure}[h]
    \centering
    \begin{minipage}{0.95\linewidth}
        \vspace{-4pt}
        \includegraphics[width=\linewidth]{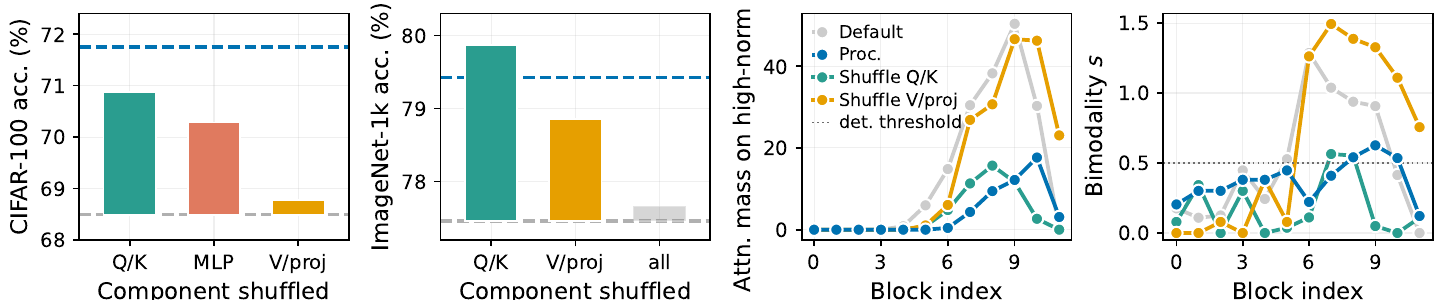}
        \vspace{-11pt}
        \makebox[\linewidth][l]{%
            \makebox[0.25\linewidth][c]{~~~~\scriptsize (a)}%
            \makebox[0.25\linewidth][c]{~~~~\scriptsize (b)}%
            \makebox[0.25\linewidth][c]{~~~~\scriptsize (c)}%
            \makebox[0.25\linewidth][c]{~~~~\scriptsize (d)}%
        }
        \vspace{-4pt}
    \end{minipage}
    \caption{\textbf{Transfer is concentrated in the value/output pathway.}
    \textbf{(a,b)} Component-shuffling accuracy.
    \textbf{(c,d)} Shuffling V/proj.\ reverts to
    the default behavior with high-norm tokens,
    while shuffling Q/K preserves the \emph{Procedural Core} effect.}
    \label{fig:where}
\end{figure}

\textbf{Transferable structure is concentrated in the value/output pathway.}
We ablate the effect of \textit{Procedural Core} on specific weight matrices
by shuffling their weights prior to image-based training.
This destroys their internal structure without
altering other properties like weight magnitudes.
Figure~\ref{fig:where} (a,b) shows that on both \textsc{CIFAR-100}
(ViT-T) and \textsc{ImageNet-1k} (ViT-B), shuffling the attention V/proj.\ matrices
causes the largest drop in performance,
whereas shuffling Q/K/MLPs preserves much of the gains.
Panels~(c,d) show the corresponding effect on token dynamics: shuffling V/proj.\
restores a default-like attention to high-norm tokens and bimodality,
while shuffling Q/K preserves the suppression. Together, these results identify the
value/output pathway as carrying much of the transferable structure responsible
for suppressing high-norm tokens.

\vspace{-2pt}
\subsection{Improved Representations Benefit a Variety of Visual Tasks}
\vspace{-1pt}
\label{sec:downstream}

\begin{figure}[t]
    \vspace{-2pt}
    \centering
    \begin{minipage}{0.95\linewidth}
        \centering
        \makebox[\linewidth][l]{%
            \hspace{0.015\linewidth}%
            \makebox[0.25\linewidth][c]{\scriptsize\textsc{ADE20K}}%
            \makebox[0.25\linewidth][c]{\scriptsize\textsc{ImageNet-S}}%
            \makebox[0.25\linewidth][c]{\scriptsize\textsc{VOC07}}%
            \makebox[0.25\linewidth][c]{\scriptsize\textsc{NYUv2}}%
        }
        \vspace{5pt}
        \includegraphics[
            width=\linewidth,
            trim=0 0 0 0,
            clip
        ]{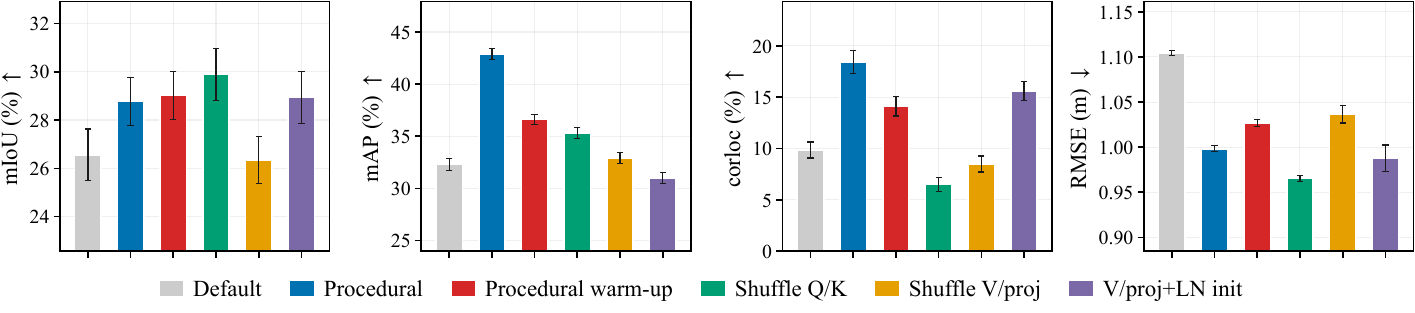}
    \end{minipage}
    \vspace{-10pt}
    \caption{\textbf{Procedural Core improves representations, particularly for dense prediction tasks} 
    (segmentation, localization, and depth estimation; frozen-backbone performance).
    This is consistent with our finding of preserving local visual information
    by suppressing high-norm tokens.
    }
    \label{fig:downstream}
    \vspace{-12pt}
\end{figure}

Prior work shows that high-norm tokens
interfere with the representation of local visual information~\citep{darcet2024visionregisters}.
Therefore, the suppression of high-norm tokens with Procedural Core
should particularly benefit tasks that require spatially local information.
Following~\citep{jiang2026visiondontneed}, we evaluate
frozen ViT-B backbones on semantic segmentation (\textsc{ADE20K}~\citep{zhou2019semantic}),
zero-shot segmentation (\textsc{ImageNet-S}~\citep{gao2022large}), unsupervised object
localization (\textsc{VOC07}~\citep{everingham2010pascal}), and monocular depth estimation
(\textsc{NYUv2}~\cite{silberman2012indoor}). Full protocols and results are provided in
Appendix~\ref{app:downstream}.

Figure~\ref{fig:downstream} shows that \emph{Procedural Core} improves performance across all four frozen-backbone evaluations, with particularly large gains on \textsc{ImageNet-S} and \textsc{VOC07}. Consistent with the component
analysis of Section~\ref{sec:downstream_analysis},
shuffling V/proj.\ removes much of the benefit, whereas
shuffling Q/K retains most of it. Conversely, initializing only V/proj.\ and
LayerNorm with \emph{Procedural Core}
(and other weights with a default random initialization)
recovers strong performance across several tasks.
\emph{Procedural Core} also matches or exceeds
the full direct warm-up on all four primary metrics.
Under the stronger DeiT-III recipe,
we observe no improvement in \textsc{ImageNet-1k} classification,
but retain gains in zero-shot segmentation and unsupervised object localization
(see Appendix~\ref{app:downstream}).

\vspace{-3pt}
\section{Discussion}
\vspace{-3pt}
We presented \emph{Procedural Core}, a generic initialization for transformers obtained by training a small recurrent model on procedural data and expanding its compact weights to models of arbitrary size. This provides a deterministic alternative to random initialization that improves supervised and self-supervised learning for vision, as well as language modeling on natural language and code.

\textbf{Condensing the benefits of abstract data.}
Our method builds on recent works
that demonstrated
the value of an initial training phase 
on abstract data
\citep{chiang2022transferability,hu2025between,jiang2026proceduralpretraining,zhang2024intelligence}.
Our contribution makes the benefits (improved convergence and generalization)
much more accessible
by dispensing with the need for a dedicated training phase.
Instead, we prepare a core set of weights
that can be reused to initialize new models of arbitrary size,
as a direct replacement for a standard random initialization.

\textbf{Recurrent transformers.}
Our method also builds on the success of looped transformers~\citep{saunshi2025reasoning,zhu2025scaling}.
In our case, the depth-wise recurrence acts as a regularizer,
forcing the model to implement generic computations in a compact set of weights.
Our analysis shows that this makes the weights 
intrinsically more reusable.

\textbf{Generic computations.}
The motivation for abstract data
is to teach 
generic mechanisms,
with training examples that avoid
the noise and biases of natural data.
Our results on multiple vision and language tasks
support the broad effects of this approach.
But prior work also shows that different types of procedural data
improve different capabilities~\citep{shinnick2025transformers}.
An important scientific question 
is to determine which properties and structures in the data
really matter, and how they lead to the acquisition of useful circuits for specific target tasks.

\textbf{Limitations \& future work.} 
Our main experiments focus on standard ViTs.
An extension to other architectures
and larger scales
is important as future work.
With the heavily optimized DeiT-III recipe
(see additional results in Appendix~\ref{app:downstream} and discussion in Appendix~\ref{app:extended_discussion}),
the gains in ImageNet classification accuracy disappear,
but the improvements persist
for segmentation and localization, consistent with
the improved representations found in
Section~\ref{sec:downstream_analysis}.
The interaction of \emph{Procedural Core} with details of the training recipe
could be studied and
possibly optimized to amplify its benefits. 
The choice of the procedural data was inherited from prior work
and could also be optimized or even substituted with mechanistic approaches to constructing model weights~\citep{giannou2023looped}.

\subsection*{AI use statement}
Generative AI was used solely to assist with grammar and presentation. It did not contribute to the development of the core ideas, experimental design or execution, or analysis
of the results.

\subsection*{Reproducibility statement}
We provide detailed descriptions of our methods, experimental settings, and evaluation procedures in the main paper and appendix. We will release code 
to reproduce our experiments upon publication.
\bibliography{main}

@String(CVPR  = {IEEE Conf. Comput. Vis. Pattern Recog. (CVPR)})

@String(ICCV  = {Int. Conf. Comput. Vis. (ICCV)})

@String(ECCV  = {Eur. Conf. Comput. Vis. (ECCV)})

@String(NeurIPS = {Adv. Neural Inform. Process. Syst. (NeurIPS)})

@String(ICML  = {Int. Conf. Mach. Learn. (ICML)})

@String(ICLR  = {Int. Conf. Learn. Represent. (ICLR)})

@String(BMVC  = {Brit. Mach. Vis. Conf. (BMVC)})

@String(AAAI  = {AAAI})

@inproceedings{lindemann2024sip,
  title        = {{SIP}: Injecting a Structural Inductive Bias into a Seq2Seq Model by Simulation},
  author       = {Lindemann, Matthias and Koller, Alexander and Titov, Ivan},
  booktitle    = {Annual Meeting of the Association for Computational Linguistics},
  year         = {2024}
}

@article{trockman2023mimetic,
  title   = {Mimetic Initialization of Self-Attention Layers},
  author  = {Trockman, Asher and Kolter, J. Zico},
  journal = {arXiv preprint arXiv:2305.09828},
  year    = {2023}
}

@article{zheng2025structured,
  title={Structured Initialization for Vision Transformers},
  author={Zheng, Jianqiao and Li, Xueqian and Saratchandran, Hemanth and Lucey, Simon},
  journal={arXiv preprint arXiv:2505.19985},
  year={2025}
}

@inproceedings{nakamura2024scaling,
  title      = {Scaling Backwards: Minimal Synthetic Pre-Training?},
  author     = {Nakamura, Ryo and Tadokoro, Ryu and Yamada, Ryosuke and Asano, Yuki M. and Laina, Iro and Rupprecht, Christian and Inoue, Nakamasa and Yokota, Rio and Kataoka, Hirokatsu},
  booktitle  = ECCV,
  year       = {2024}
}

@article{wu2022insights,
  title   = {Insights into Pre-Training via Simpler Synthetic Tasks},
  author  = {Wu, Yuhuai and Li, Felix and Liang, Percy S.},
  journal = NeurIPS,
  year    = {2022}
}

@article{mccoy2023modeling,
  title   = {Modeling Rapid Language Learning by Distilling Bayesian Priors into Artificial Neural Networks},
  author  = {McCoy, R. Thomas and Griffiths, Thomas L.},
  journal = {arXiv preprint arXiv:2305.14701},
  year    = {2023}
}

@article{papadimitriou2023injecting,
  title   = {Injecting Structural Hints: Using Language Models to Study Inductive Biases in Language Learning},
  author  = {Papadimitriou, Isabel and Jurafsky, Dan},
  journal = {arXiv preprint arXiv:2304.13060},
  year    = {2023}
}

@article{zhang2024intelligence,
  title   = {Intelligence at the Edge of Chaos},
  author  = {Zhang, Shiyang and Patel, Aakash and Rizvi, Syed A. and Liu, Nianchen and He, Sizhuang and Karbasi, Amin and Zappala, Emanuele and van Dijk, David},
  journal = {arXiv preprint arXiv:2410.02536},
  year    = {2024}
}

@inproceedings{chiang2022transferability,
  title      = {On the Transferability of Pre-Trained Language Models: A Study from Artificial Datasets},
  author     = {Chiang, Cheng-Han and Lee, Hung-Yi},
  booktitle  = AAAI,
  year       = {2022}
}

@inproceedings{hu2025between,
  title     = {Between Circuits and Chomsky: Pre-pretraining on Formal Languages Imparts Linguistic Biases},
  author    = {Hu, Michael Y. and Petty, Jackson and Shi, Chuan and Merrill, William and Linzen, Tal},
  booktitle = {Annual Meeting of the Association for Computational Linguistics},
  year      = {2025}
}

@article{radford2019language,
  title   = {Language Models Are Unsupervised Multitask Learners},
  author  = {Radford, Alec and Wu, Jeffrey and Child, Rewon and Luan, David and Amodei, Dario and Sutskever, Ilya and others},
  journal = {OpenAI Blog},
  year    = {2019}
}

@inproceedings{huang2020improving,
  title      = {Improving Transformer Optimization Through Better Initialization},
  author     = {Huang, Xiao Shi and Perez, Felipe and Ba, Jimmy and Volkovs, Maksims},
  booktitle  = ICML,
  year       = {2020}
}

@article{xu2023initializing,
  title   = {Initializing Models with Larger Ones},
  author  = {Xu, Zhiqiu and Chen, Yanjie and Vishniakov, Kirill and Yin, Yida and Shen, Zhiqiang and Darrell, Trevor and Liu, Lingjie and Liu, Zhuang},
  journal = {arXiv preprint arXiv:2311.18823},
  year    = {2023}
}

@article{han2025learning,
  title={Learning to See Before Seeing: Demystifying LLM Visual Priors from Language Pre-training},
  author={Han, Junlin and Tong, Shengbang and Fan, David and Ren, Yufan and Sinha, Koustuv and Torr, Philip and Kokkinos, Filippos},
  journal={arXiv preprint arXiv:2509.26625},
  year={2025}
}

@inproceedings{baradad2022learninglookingnoise,
  title      = {Learning to See by Looking at Noise},
  author     = {Baradad, Manel and Wulff, Jonas and Wang, Tongzhou and Isola, Phillip and Torralba, Antonio},
  booktitle  = NeurIPS,
  year       = {2021}
}

@inproceedings{kataoka2022replacinglabeledrealimagedatasets,
  title      = {Replacing Labeled Real-Image Datasets with Auto-Generated Contours},
  author     = {Kataoka, Hirokatsu and Hayamizu, Ryo and Yamada, Ryosuke and Nakashima, Kodai and Takashima, Sora and Zhang, Xinyu and Martinez-Noriega, Edgar Josafat and Inoue, Nakamasa and Yokota, Rio},
  booktitle  = CVPR,
  year       = {2022}
}

@inproceedings{nakamura2023pretrainingvisiontransformerslimited,
  title      = {Pre-Training Vision Transformers with Very Limited Synthesized Images},
  author     = {Nakamura, Ryo and Kataoka, Hirokatsu and Takashima, Sora and Martinez-Noriega, Edgar Josafat and Yokota, Rio and Inoue, Nakamasa},
  booktitle  = ICCV,
  year       = {2023}
}

@article{zhao2022zero,
  title   = {ZerO Initialization: Initializing Neural Networks with Only Zeros and Ones},
  author  = {Zhao, Jiawei and Schäfer, Florian and Anandkumar, Anima},
  journal = {arXiv preprint arXiv:2110.12661},
  year    = {2021}
}

@article{giri2025ibitutilizinginductivebiases,
  title   = {{IBiT}: Utilizing Inductive Biases to Create a More Data Efficient Attention Mechanism},
  author  = {Giri, Adithya},
  journal = {arXiv preprint arXiv:2509.22719},
  year    = {2025}
}

@inproceedings{huh2024position,
  title      = {Position: The Platonic Representation Hypothesis},
  author     = {Huh, Minyoung and Cheung, Brian and Wang, Tongzhou and Isola, Phillip},
  booktitle  = ICML,
  year       = {2024}
}

@inproceedings{shinnick2025transformers,
  title={Transformers Pretrained on Procedural Data Contain Modular Structures for Algorithmic Reasoning},
  author={Shinnick, Zachary and Jiang, Liangze and Saratchandran, Hemanth and van den Hengel, Anton and Teney, Damien},
  booktitle={ICML Workshop on Methods and Opportunities at Small Scale},
  year={2025}
}

@article{du2025large,
  title={Large language model for lossless image compression with visual prompts},
  author={Du, Junhao and Zhou, Chuqin and Cao, Ning and Chen, Gang and Chen, Yunuo and Cheng, Zhengxue and Song, Li and Lu, Guo and Zhang, Wenjun},
  journal={arXiv preprint arXiv:2502.16163},
  year={2025}
}

@article{zheng2024lm4lv,
  title={Lm4lv: A frozen large language model for low-level vision tasks},
  author={Zheng, Boyang and Gu, Jinjin and Li, Shijun and Dong, Chao},
  journal={arXiv preprint arXiv:2405.15734},
  year={2024}
}

@article{dosovitskiy2020image,
  title={An image is worth 16x16 words: Transformers for image recognition at scale},
  author={Dosovitskiy, Alexey},
  journal={arXiv preprint arXiv:2010.11929},
  year={2020}
}

@inproceedings{deng2009imagenet,
  title        = {ImageNet: A Large-Scale Hierarchical Image Database},
  author       = {Deng, Jia and Dong, Wei and Socher, Richard and Li, Li-Jia and Li, Kai and Fei-Fei, Li},
  booktitle    = CVPR,
  year         = {2009}
}

@inproceedings{shinnick2025learnwithoutimages,
    author = {Shinnick, Zachary and Jiang, Liangze and Saratchandran, Hemanth and Teney, Damien and van den Hengel, Anton},
    title = {Can You Learn to See Without Images? Procedural Warm-Up for Vision Transformers},
    booktitle = CVPR,
    year = {2026}
}

@article{jiang2026proceduralpretraining,
  title   = {Procedural Pretraining: Warming Up Language Models with Abstract Data},
  author  = {Jiang, Liangze and Shinnick, Zachary and van den Hengel, Anton and Saratchandran, Hemanth and Teney, Damien},
  journal = {arXiv preprint arXiv:2601.21725},
  year    = {2026},
}

@inproceedings{caron2021emerging,
  title={Emerging Properties in Self-Supervised Vision Transformers},
  author={Caron, Mathilde and Touvron, Hugo and Misra, Ishan and J\'egou, Herv\'e  and Mairal, Julien and Bojanowski, Piotr and Joulin, Armand},
  booktitle={International Conference on Computer Vision (ICCV)},
  year={2021}
}

@inproceedings{lan2020albert,
      title={ALBERT: A Lite BERT for Self-supervised Learning of Language Representations}, 
      author={Lan, Zhenzhong and Chen, Mingda and Goodman, Sebastian and Gimpel, Kevin and Sharma, Piyush and Soricut, Rad},
      year={2020},
      booktitle=ICLR
}

@inproceedings{dehghani2019universaltransformers,
      title={Universal Transformers}, 
      author={Dehghani, Mostafa and Gouws, Stephan and Vinyals, Oriol and Uszkoreit, Jakob and Kaiser, Lukasz},
      year={2019},
      booktitle=ICLR
}

@article{saunshi2024inductivebiasstackingimproving,
      title={On the Inductive Bias of Stacking Towards Improving Reasoning}, 
      author={Saunshi, Nikunj and Karp, Stefani  and Krishnan, Shankar and Miryoosefi, Sobhan and Reddi, Sashank J. and Kumar, Sanjiv},
      year={2024},
      journal={arXiv preprint arXiv:2409.19044}
}

@inproceedings{jacobs2025block,
  title={Block-Recurrent Dynamics in Vision Transformers},
  author={Jacobs, Mozes and Fel, Thomas and Hakim, Richard and Brondetta, Alessandra and Ba, Demba and Keller, T Andy},
  journal=ICLR,
  year={2025}
}

@inproceedings{feng2025wave,
  title={WAVE: Weight Templates for Adaptive Initialization of Variable-sized Models},
  author={Feng, Fu and Xie, Yucheng and Wang, Jing and Geng, Xin},
  booktitle=CVPR,
  year={2025}
}

@InProceedings{glorot2010understandingthedif,
  title = 	 {Understanding the difficulty of training deep feedforward neural networks},
  author = 	 {Glorot, Xavier and Bengio, Yoshua},
  booktitle = 	 {International Conference on Artificial Intelligence and Statistics},
  year = 	 {2010},

}

@inproceedings{he2016deepresiduallearning,
  author={He, Kaiming and Zhang, Xiangyu and Ren, Shaoqing and Sun, Jian},
  booktitle=CVPR, 
  title={Deep Residual Learning for Image Recognition}, 
  year={2016},
}

@inproceedings{chen2022bert2bert,
  title={bert2bert: Towards reusable pretrained language models},
  author={Chen, Cheng and Yin, Yichun and Shang, Lifeng and Jiang, Xin and Qin, Yujia and Wang, Fengyu and Wang, Zhi and Chen, Xiao and Liu, Zhiyuan and Liu, Qun},
  booktitle={Annual Meeting of the Association for Computational Linguistics (Volume 1: Long Papers)},
  year={2022}
}

@article{samragh2024scaling,
  title={Scaling smart: Accelerating large language model pre-training with small model initialization},
  author={Samragh, Mohammad and Mirzadeh, Iman and Vahid, Keivan Alizadeh and Faghri, Fartash and Cho, Minsik and Nabi, Moin and Naik, Devang and Farajtabar, Mehrdad},
  journal={arXiv preprint arXiv:2409.12903},
  year={2024}
}

@article{panigrahi2024efficient,
  title={Efficient stagewise pretraining via progressive subnetworks},
  author={Panigrahi, Abhishek and Saunshi, Nikunj and Lyu, Kaifeng and Miryoosefi, Sobhan and Reddi, Sashank and Kale, Satyen and Kumar, Sanjiv},
  journal={arXiv preprint arXiv:2402.05913},
  year={2024}
}

@article{saunshi2025reasoning,
  title={Reasoning with latent thoughts: On the power of looped transformers},
  author={Saunshi, Nikunj and Dikkala, Nishanth and Li, Zhiyuan and Kumar, Sanjiv and Reddi, Sashank J},
  journal={arXiv preprint arXiv:2502.17416},
  year={2025}
}

@article{zhu2025scaling,
  title={Scaling latent reasoning via looped language models},
  author={Zhu, Rui-Jie and Wang, Zixuan and Hua, Kai and Zhang, Tianyu and Li, Ziniu and Que, Haoran and Wei, Boyi and Wen, Zixin and Yin, Fan and Xing, He and others},
  journal={arXiv preprint arXiv:2510.25741},
  year={2025}
}

@article{buzeta2026seeing,
  title={Seeing to Generalize: How Visual Data Corrects Binding Shortcuts},
  author={Buzeta, Nicolas and del Rio, Felipe and Hinostroza, Cristian and Parra, Denis and Lobel, Hans and Icarte, Rodrigo Toro},
  journal={arXiv preprint arXiv:2602.15183},
  year={2026}
}

@article{riachi2025random,
  title={Random Initialization Can't Catch Up: The Advantage of Language Model Transfer for Time Series Forecasting},
  author={Riachi, Roland and Rasul, Kashif and Ashok, Arjun and Humane, Prateek and Roger, Alexis and Williams, Andrew R and Nevmyvaka, Yuriy and Rish, Irina},
  journal={arXiv preprint arXiv:2506.21570},
  year={2025}
}

@article{gupta2025better,
  title={Better together: Leveraging unpaired multimodal data for stronger unimodal models},
  author={Gupta, Sharut and Sundaram, Shobhita and Wang, Chenyu and Jegelka, Stefanie and Isola, Phillip},
  journal={arXiv preprint arXiv:2510.08492},
  year={2025}
}

@inproceedings{lu2022frozen,
  title={Frozen pretrained transformers as universal computation engines},
  author={Lu, Kevin and Grover, Aditya and Abbeel, Pieter and Mordatch, Igor},
  booktitle=AAAI,
  year={2022}
}

@inproceedings{
penedo2024the,
title={The FineWeb Datasets: Decanting the Web for the Finest Text Data at Scale},
author={Guilherme Penedo and Hynek Kydl{\'\i}{\v{c}}ek and Loubna Ben allal and Anton Lozhkov and Margaret Mitchell and Colin Raffel and Leandro Von Werra and Thomas Wolf},
booktitle=NeurIPS,
year={2024},
}

@misc{codeparrot,
  author = {HuggingFace},
  title = {{CodeParrot} Dataset Cleaned},
  year = {2022},
}

@inproceedings{giannou2023looped,
  title={Looped transformers as programmable computers},
  author={Giannou, Angeliki and Rajput, Shashank and Sohn, Jy-yong and Lee, Kangwook and Lee, Jason D and Papailiopoulos, Dimitris},
  booktitle=ICML,
  year={2023},
  organization={PMLR}
}

@article{mcleish2025teaching,
  title={Teaching Pretrained Language Models to Think Deeper with Retrofitted Recurrence},
  author={McLeish, Sean and Li, Ang and Kirchenbauer, John and Kalra, Dayal Singh and Bartoldson, Brian R and Kailkhura, Bhavya and Schwarzschild, Avi and Geiping, Jonas and Goldstein, Tom and Goldblum, Micah},
  journal={arXiv preprint arXiv:2511.07384},
  year={2025}
}

@article{skean2025layer,
  title={Layer by layer: Uncovering hidden representations in language models},
  author={Skean, Oscar and Arefin, Md Rifat and Zhao, Dan and Patel, Niket and Naghiyev, Jalal and LeCun, Yann and Shwartz-Ziv, Ravid},
  journal={arXiv preprint arXiv:2502.02013},
  year={2025}
}

@article{hoffmann2022training,
  title={Training compute-optimal large language models},
  author={Hoffmann, Jordan and Borgeaud, Sebastian and Mensch, Arthur and Buchatskaya, Elena and Cai, Trevor and Rutherford, Eliza and Casas, DDL and Hendricks, Lisa Anne and Welbl, Johannes and Clark, Aidan and others},
  journal={arXiv preprint arXiv:2203.15556},
  volume={10},
  year={2022}
}

@article{jiang2026visiondontneed,
  title={Vision Transformers Don't Need Trained Registers},
  author={Jiang, Nicholas and Dravid, Amil and Efros, Alexei and Gandelsman, Yossi},
  journal=NeurIPS,
  year={2026}
}

@inproceedings{darcet2024visionregisters,
  title={Vision transformers need registers},
  author={Darcet, Timoth{\'e}e and Oquab, Maxime and Mairal, Julien and Bojanowski, Piotr},
  booktitle=ICLR,
  year={2024}
}

@inproceedings{touvron2022deit,
  title={{DeiT III}: Revenge of the {ViT}},
  author={Touvron, Hugo and Cord, Matthieu and J{\'e}gou, Herv{\'e}},
  booktitle=ECCV,
  year={2022},
  }

@article{zhou2019semantic,
  title={Semantic understanding of scenes through the ade20k dataset},
  author={Zhou, Bolei and Zhao, Hang and Puig, Xavier and Xiao, Tete and Fidler, Sanja and Barriuso, Adela and Torralba, Antonio},
  journal={International journal of computer vision},
  year={2019},
  }

@article{gao2022large,
  title={Large-scale unsupervised semantic segmentation},
  author={Gao, Shanghua and Li, Zhong-Yu and Yang, Ming-Hsuan and Cheng, Ming-Ming and Han, Junwei and Torr, Philip},
  journal={IEEE transactions on pattern analysis and machine intelligence},
  year={2022},
  }

@article{everingham2010pascal,
  title={The Pascal Visual Object Classes (VOC) Challenge},
  author={Everingham, Mark and Van Gool, Luc and Williams, Christopher K. I. and Winn, John and Zisserman, Andrew},
  journal={International Journal of Computer Vision},
  year={2010},
  }

@inproceedings{silberman2012indoor,
  title={Indoor segmentation and support inference from rgbd images},
  author={Silberman, Nathan and Hoiem, Derek and Kohli, Pushmeet and Fergus, Rob},
  booktitle=ECCV,
  year={2012},
}

@inproceedings{simeoni2021localizing,
  title={Localizing Objects with Self-Supervised Transformers and no Labels},
  author={Sim{\'e}oni, Oriane and Puy, Gilles and Vo, Huy V. and Roburin, Simon
          and Gidaris, Spyros and Bursuc, Andrei and P{\'e}rez, Patrick
          and Marlet, Renaud and Ponce, Jean},
  booktitle=BMVC,
  year={2021}
}
\bibliographystyle{iclr2027_conference}

\clearpage
\appendix


\section{Training Details}

\subsection{Procedural Core Training Setup}
\label{sec:pwu}

To obtain the Procedural Core, we follow the procedural warm-up framework
introduced in~\citep{shinnick2025learnwithoutimages}.
We use a ViT-T/16 backbone as the auxiliary model, trained for 15{,}000 steps
using masked-token prediction on procedurally generated $k$-Dyck sequences.
All procedural training was performed on a single NVIDIA H200 GPU.

\paragraph{Recurrent parameterization.}
The auxiliary model is structured with a distinct input layer (layer 0) and
output layer (layer 11), while intermediate layers (1--10) share parameters.

\paragraph{Cross-instance parameterization.}
By default, we share parameters across $n=3$ parallel instances trained with
different random seeds. Empirically, cross-instance sharing provides only
marginal and inconsistent gains. Attention and MLP weights are shared across
instances, while LayerNorm and head parameters remain independent.

\begin{table}[h]
\centering
\small
\caption{Optimization settings used to train the auxiliary model for learning
the Procedural Core.}
\label{tab:train-hparams}
\begin{tabular}{l l}
\toprule
\textbf{Setting} & \textbf{Value} \\
\midrule
Batch size & 256 \\
Training steps & 15{,}000 \\
Mask ratio & 0.5 (close-only) \\
Optimizer & AdamW \\
Learning rate & $2 \times 10^{-3}$ \\
Weight decay & 0.05 \\
Betas & (0.9, 0.999) \\
LR schedule & Cosine decay \\
Warmup steps & 1{,}000 \\
\bottomrule
\end{tabular}
\end{table}

\paragraph{Procedural data generation.}
We generate hierarchical procedural sequences
following~\citep{shinnick2025learnwithoutimages}.
Opening tokens are drawn from 64 distinct types and closing tokens from
64 corresponding types. Whenever structurally permissible, an opening token
is sampled with probability $p_{\text{open}} = 0.6$.

\subsection{Supervised Image Classification}
\label{sec:abs_training_detail_supervised}

For \textsc{ImageNet-1K} training (Figure~\ref{fig:imagenet_results}), we
adopt the standard ViT-B training protocol of~\citep{xu2023initializing}.
Optimization uses AdamW with cosine learning-rate decay and linear warmup,
together with RandAugment, Mixup, CutMix, and label smoothing. We train for
300 epochs using a base learning rate of $2 \times 10^{-3}$, a batch size of
4096, and a 50-epoch linear warm-up. All supervised \textsc{ImageNet-1K}
experiments were conducted on either 8 NVIDIA H200 GPUs or 4 NVIDIA A100 GPUs,
depending on availability. The training configuration and effective batch size
were kept identical across runs.

For the fine-tuning experiment in Table~\ref{tab:cifar_finetune}, we initialize
from the \textsc{ImageNet-1K} ViT-B checkpoints and fine-tune on
\textsc{CIFAR-100}. We follow the same optimization protocol as above, but
reduce the batch size to 512 and train for 50 epochs with 5 warmup epochs and
a base learning rate of $5 \times 10^{-4}$.

\subsection{Self-Supervised Training (DINO)}
\label{sec:abs_training_detail_selfsupervised}

We adopt the DINO framework~\citep{caron2021emerging} with a ViT-S/16
backbone. Training follows the standard configuration with multi-crop
augmentation and momentum teacher updates, using AdamW with cosine
learning-rate decay for 300 epochs (10 warmup epochs). The base learning rate
is $5 \times 10^{-4}$, and the batch size is 64 per GPU across 4 NVIDIA A100
GPUs (total batch size 256).

We evaluate representation quality using the standard DINO weighted k-NN
protocol, where L2-normalized ViT-S/16 features are compared using cosine
similarity and each validation image is classified by a temperature-scaled
vote over its $k=20$ nearest neighbors (temperature $\tau=0.07$).
We additionally report linear evaluation results using the standard DINO
protocol with a linear classifier trained on frozen backbone features.

\subsection{Procedural Core for Language Models}
\label{sec:lm_details}

To obtain the Procedural Core for language modeling, we autoregressively train
a GPT-2--style transformer (12 layers, hidden size 768, 12 attention heads) on
procedurally generated $k$-Dyck sequences, using $k=64$ opening and closing
symbols and sequence length 2048.

Training is performed for 2{,}000 steps using AdamW with cosine learning-rate
decay from $5 \times 10^{-4}$, including 200 warmup steps, and weight decay
0.1. The effective batch size is 32 sequences per update, corresponding to
approximately 131M tokens of procedural training. Training is performed on a
single NVIDIA RTX 4090 GPU.

We use the same recurrent auxiliary architecture described in
Appendix~\ref{sec:pwu}, where the input and output layers are distinct and the
intermediate transformer blocks share parameters.

\subsection{Language Modeling Training Details}

We train GPT-2-style models (12 layers, 768 hidden size, 12 heads) with
context length 1024 on both \textsc{FineWeb-Edu} and \textsc{CodeParrot}.
Training is performed for a total of 2B tokens. For \textsc{CodeParrot},
sequences are tokenized using the CodeParrot tokenizer and sampled from the
\texttt{codeparrot-clean} splits. All other hyperparameters are listed in
Table~\ref{tab:lm_hparams}. All language training experiments were conducted
on a single NVIDIA RTX 4090 GPU.

\begin{table}[h]
\centering
\small
\caption{GPT-2 Small training configuration for \textsc{FineWeb-Edu} and
\textsc{CodeParrot} experiments.}
\label{tab:lm_hparams}
\begin{tabular}{l l}
\toprule
\textbf{Setting} & \textbf{Value} \\
\midrule
Architecture & GPT-2 Small (12L, 768D, 12H) \\
Context length & 1024 \\
Training budget & 2B tokens \\
Effective batch size & 32 sequences (32{,}768 tokens) \\
Optimizer & AdamW \\
Learning rate & $6\times10^{-4}$ (cosine to $6\times10^{-5}$) \\
Warmup & 5\% of total steps \\
Weight decay & 0.1 \\
Gradient clipping & 1.0 \\
\bottomrule
\end{tabular}
\end{table}


\section{Additional Experimental Results}


\subsection{Comparison with Procedural Warm-Up}
\label{app:warmup-comparison}

We additionally compare \emph{Procedural Core} with standard procedural
warm-up~\citep{shinnick2025learnwithoutimages} on \textsc{CIFAR-100}.
Using the same ViT-Tiny target architecture, \emph{Procedural Core}
improves top-1 accuracy from $73.9\%$ to $74.6\%$
(Table~\ref{tab:warmup-cifar}), despite avoiding procedural training of the
target model itself.

\begin{table}[t]
    \centering
    \small
    \caption{\textbf{Comparison with procedural warm-up on
    \textsc{CIFAR-100}.}
    ViT-Tiny top-1 accuracy (mean $\pm$ std).}
    \label{tab:warmup-cifar}
    \begin{tabular}{lc}
        \toprule
        & Top-1 Acc. (\%) \\
        \midrule
        Procedural warm-up & $73.9 \pm 0.1$ \\
        \addlinespace[2pt]
        \cellcolor{coreblue}\textbf{Procedural Core (ours)}
        & \cellcolor{coreblue}\textbf{$74.6 \pm 0.5$} \\
        \bottomrule
    \end{tabular}
\end{table}


\subsection{Downstream Transfer after ImageNet Pretraining}
\label{app:cifar-transfer}

We fine-tune the \textsc{ImageNet-1K}-pretrained ViT-Base checkpoints on
\textsc{CIFAR-100} using an identical protocol across initializations
(Section~\ref{sec:abs_training_detail_supervised}). As shown in
Table~\ref{tab:cifar_finetune}, \emph{Procedural Core} achieves the
highest downstream accuracy, indicating that its benefit persists after
large-scale \textsc{ImageNet-1K} training.

\begin{table}[t]
    \centering
    \caption{\textbf{Downstream transfer on \textsc{CIFAR-100}.}
    Fine-tuning accuracy after \textsc{ImageNet-1K} pretraining
    (mean $\pm$ std over three seeds).}
    \label{tab:cifar_finetune}

    {\small
    \begin{tabular}{lc}
        \toprule
        Initialization & Top-1 (\%) \\
        \midrule
        Default init. & $88.4 \pm 0.8$ \\
        Mimetic init. & $89.1 \pm 0.5$ \\
        Procedural warm-up & $89.4 \pm 0.3$ \\
        \addlinespace[2pt]
        \cellcolor{coreblue}\textbf{Procedural Core}
        & \cellcolor{coreblue}\textbf{$89.7 \pm 0.1$} \\
        \bottomrule
    \end{tabular}
    }
\end{table}


\subsection{Linear Evaluation of DINO Representations}
\label{sec:linear-probe}

Linear-probe results (Table~\ref{tab:dino_linear_probe}) show only marginal
differences compared with k-NN evaluation. This is expected because a
trainable linear classifier can partially compensate for weaker
representations, whereas k-NN more directly reflects intrinsic
representation quality.

\begin{table}[t]
\centering
\caption{\textbf{Linear evaluation of DINO representations on
\textsc{ImageNet-1K}.} Frozen ViT-S features evaluated with a linear
classifier.}
\label{tab:dino_linear_probe}
\begin{scriptsize}
\begin{tabular}{l c}
\toprule
Initialization & Top-1 (\%) \\
\midrule
Default initialization & 75.36 \\
\textbf{Procedural Core (ours)} & \textbf{75.42} \\
\bottomrule
\end{tabular}
\end{scriptsize}
\end{table}


\subsection{Spectral Analysis Details}
\label{app:spectral}
\label{sec:singular-value}

For each weight matrix $W \in \mathbb{R}^{m \times n}$, we compute its
singular value decomposition $W = U\Sigma V^\top$, with singular values
$\{\sigma_i\}$ ordered by decreasing magnitude. We define the cumulative
spectral energy captured by the top $k$ singular directions as
\begin{equation}
E(k) =
\frac{\sum_{i=1}^{k}\sigma_i^2}
{\sum_{i=1}^{\min(m,n)}\sigma_i^2}.
\end{equation}
Slower saturation of $E(k)$ indicates that spectral energy is distributed
across more singular directions. In Figure~\ref{fig:sv-energy} of the main
text, we additionally report the number of singular directions required to
capture $90\%$ and $95\%$ of the total spectral energy.

\begin{figure*}[t]
    \centering
    \includegraphics[width=\textwidth]{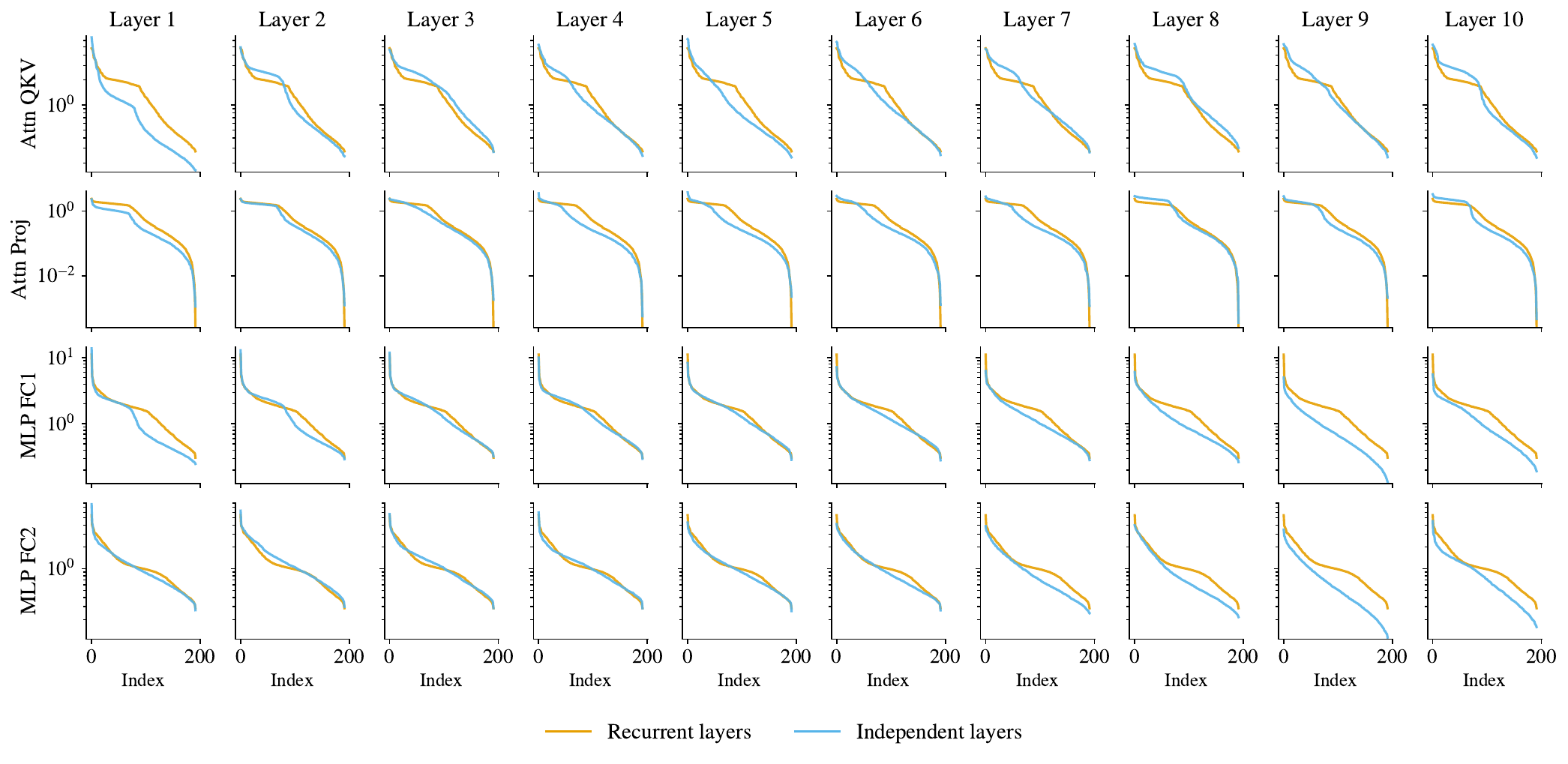}
    \caption{\textbf{Layer-wise singular value spectra under recurrent vs.\
    independent parameterization.}
    Singular value decay for attention and MLP weight matrices across
    layers 1--10. In the recurrent model, the middle layers share
    parameters, so the same spectrum appears across layers. Each subplot
    contrasts this with independently parameterized layers. Recurrent
    layers exhibit slower spectral decay, indicating a more distributed
    use of singular directions.}
    \label{fig:sv_layers_fullwidth}
\end{figure*}

Figure~\ref{fig:sv_layers_fullwidth} reports the corresponding spectra
across layers. Across attention (QKV and projection) and MLP (FC1 and FC2)
components, recurrently parameterized layers exhibit slower spectral decay
than independently parameterized baselines, showing that the effect is
consistent across depth.

\subsection{Qualitative Token-Norm Analysis}
\label{app:qualitative-token-norms}

\begin{figure}[t]
    \centering
    \includegraphics[width=1.0\linewidth]{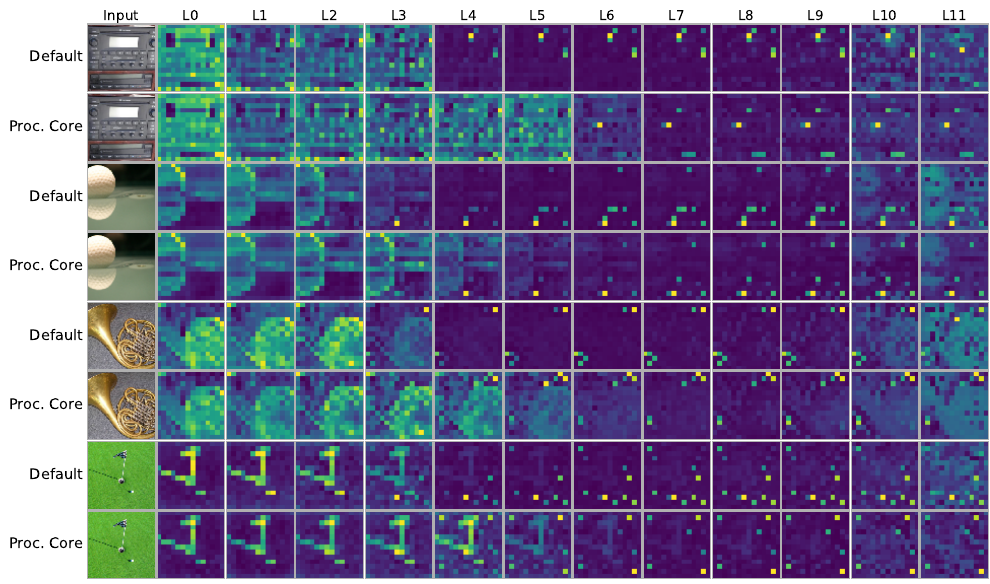}
    \caption{\textbf{Additional token-norm dynamics across depth.}
    Per-token hidden-state norms across transformer blocks for default and
    \emph{Procedural Core} initializations. The models begin with similar
    token-norm structure but increasingly diverge across depth, with the
    \emph{Procedural Core} suppressing high-norm token outliers in later
    blocks.}
    \label{fig:norm_maps_extended}
\end{figure}

Figure~\ref{fig:norm_maps_extended} provides additional qualitative
examples of the token-norm dynamics discussed in the main text.

\subsection{Token-Norm Analysis Details}
\label{app:tokennorm}

All statistics use $200$ \textsc{ImageNet-1K} validation images, sampled
once with a fixed seed and held constant across models. For each model and
block, we pool the $\ell_2$ norms of all patch-token hidden states over
images, excluding zero norms. The \textsc{cls} token is excluded throughout
as both query and key. Algorithm~\ref{alg:bimod} gives the detection
procedure, applied independently at each block.

\begin{algorithm}[H]
\caption{High-norm token detection at one block}
\label{alg:bimod}
\begin{algorithmic}[1]
\REQUIRE pooled token norms $\{n_i\}$; bins $B = 200$; smoothing window $w = 5$
\ENSURE bimodality $s$; cutoff $c$ (or none)
\STATE $h \gets$ histogram of $\{\log_{10} n_i\}$ with $B$ equal-width bins
\STATE $h \gets$ moving average of $h$ with window $w$
\STATE $p_1 \gets \arg\max h$ \hfill $\triangleright$ main mode
\STATE $v \gets \arg\min h_{[p_1,\;p_1 + 0.9(B-p_1)]}$ \hfill $\triangleright$ valley
\STATE $p_2 \gets \arg\max h_{[v,\;B]}$ \hfill $\triangleright$ second mode
\STATE $s \gets \log_{10}\!\left(\max(h_{p_2},0.5) \,/\, \max(h_v,0.5)\right)$
\IF{$p_1$ or $v$ lies within $10$ bins of the right edge}
    \RETURN $s$, none
\ENDIF
\STATE $c \gets 10^{e_v}$, where $e_v$ is the left edge of bin $v$
\STATE $f \gets$ fraction of tokens with $n_i > c$
\IF{$s \geq 0.5$ \AND $f \in [0.1\%,20\%]$}
    \RETURN $s$, $c$
\ELSE
    \RETURN $s$, none
\ENDIF
\end{algorithmic}
\end{algorithm}

The statistic $s$ measures, in orders of magnitude, how far the second mode
rises above the valley separating it from the main mode. The threshold
$s \geq 0.5$ requires approximately a factor-of-three separation, while the
constraint on $f$ rejects spurious detections arising from smooth heavy
tails.

\paragraph{Attention mass on high-norm tokens.}
This measurement is independent of the detection procedure above. At each
block, we mark tokens whose norm exceeds three times the per-image median
and sum the post-softmax attention mass they receive as keys from
patch-token queries. We normalize by total attention mass and average over
heads, queries, and images. The criterion is evaluated independently at
every block and is therefore not tied to the cutoff $c$.

\paragraph{Component shuffling.}
The interventions in Section~\ref{sec:downstream_analysis} independently
permute the entries of one component of the initialization (Q/K, V/proj, or
MLP) within each weight matrix before downstream training. This preserves
marginal weight statistics while destroying learned structure; all other
parameters remain unchanged.


\subsection{Downstream Visual Evaluation}
\label{app:downstream}

\paragraph{Evaluation protocols.}
All backbones are frozen so that differences reflect the learned
representation rather than further backbone training. On \textsc{ADE20K},
we train a $1\times1$ convolution on last-block patch tokens over the full
$20{,}210$-image training split for $10$ epochs at $448$\,px and report
mIoU on the $2{,}000$ validation images. On \textsc{ImageNet-S}, no
parameters are trained: we threshold the \textsc{cls}-to-patch attention
map at its per-image mean and score it against the ground-truth mask over
all $4{,}276$ images, reporting threshold-free mAP. On \textsc{VOC07}, we
run LOST~\citep{simeoni2021localizing} on frozen value features over all $5{,}011$ trainval
images and report corloc. On \textsc{NYUv2}, we fit a linear log-depth head
on the canonical $795/645$ split and report RMSE. Uncertainties are $95\%$
bootstrap intervals over evaluation images unless otherwise stated.

\begin{figure}[t]
\centering
\includegraphics[width=0.95\linewidth]{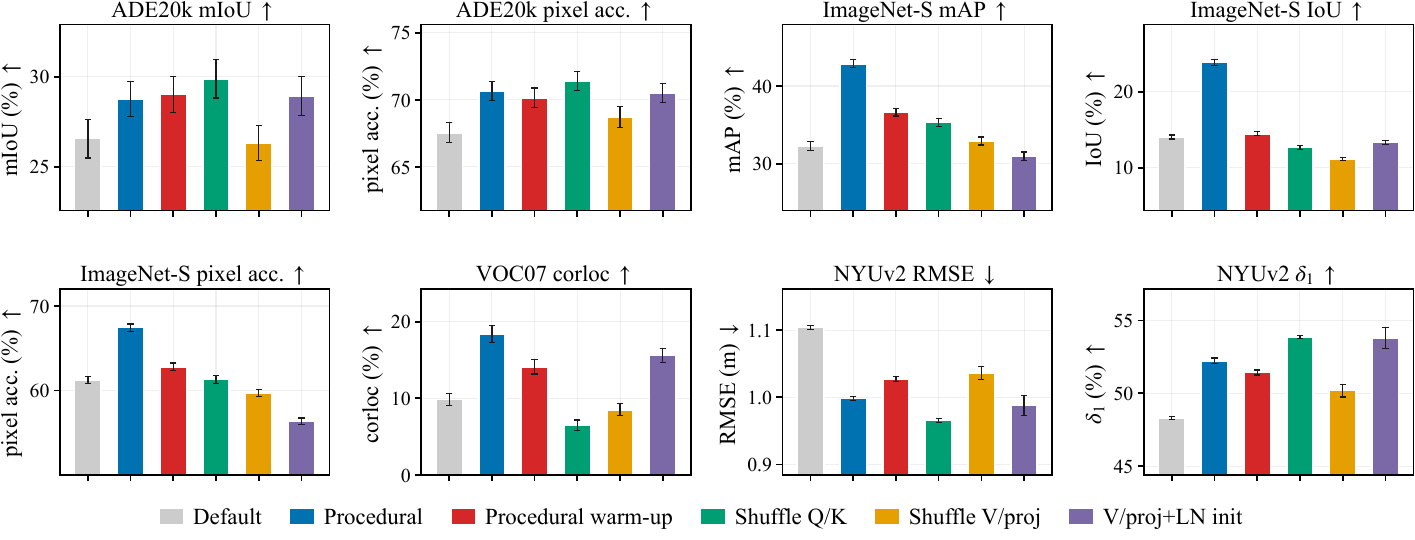}
\caption{\textbf{Full downstream evaluation across dense and localization
tasks.}
Frozen ViT-B backbones evaluated on segmentation
(\textsc{ADE20K}, \textsc{ImageNet-S}), object localization
(\textsc{VOC07}), and depth estimation (\textsc{NYUv2}). The
\emph{Procedural Core} improves over default initialization across the
primary metrics; shuffling V/proj largely removes these gains, while
transplanting V/proj+LN recovers much of them. Error bars denote $95\%$
bootstrap intervals (\textsc{NYUv2}: s.d.\ over three readout seeds).}
\label{fig:downstream-full}
\end{figure}

Figure~\ref{fig:downstream-full} reports all downstream metrics,
complementing the primary metrics presented in Figure~\ref{fig:downstream}
of the main text.

\paragraph{Component interventions.}
Shuffling V/proj removes much of the benefit of \emph{Procedural Core}.
It falls below the default baseline on \textsc{ADE20K} and \textsc{VOC07}
($26.34$ mIoU and $8.50$ corloc versus $26.56$ and $9.86$).
Transplanting only V/proj and LayerNorm recovers much of the improvement on
\textsc{ADE20K}, \textsc{VOC07}, and \textsc{NYUv2}, although not on
\textsc{ImageNet-S}. Shuffling Q/K largely preserves performance and is the
strongest arm on \textsc{ADE20K} and \textsc{NYUv2}, but is weaker on
\textsc{VOC07}, where the readout depends directly on feature geometry.

\paragraph{Aggregate comparison.}
Across all eight downstream metrics, a Friedman test rejects the null of
equal ranks across the six arms ($\chi^2=18.14$, $p=0.003$). The
\emph{Procedural Core} achieves the best mean rank ($2.00$), and a Nemenyi
post-hoc test at $\alpha=0.05$ separates it from both default initialization
($4.88$) and the V/proj-shuffled ablation ($5.12$).

\begin{figure}[t]
\centering
\includegraphics[width=0.9\linewidth]{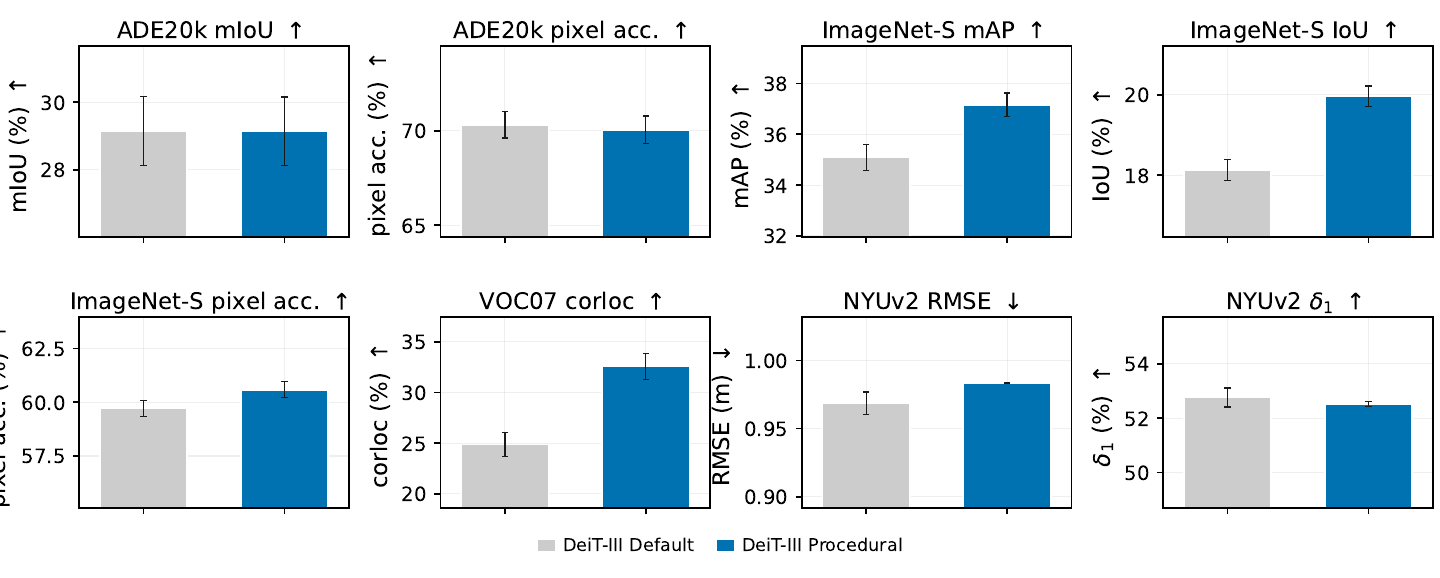}
\caption{\textbf{Under the stronger DeiT-III recipe, gains concentrate on
attention-based tasks.}
Full downstream evaluation under DeiT-III~\citep{touvron2022deit}.
\emph{Procedural Core} improves zero-shot segmentation on
\textsc{ImageNet-S} and unsupervised object localization on \textsc{VOC07},
while performance is broadly unchanged on the remaining tasks.}
\label{fig:deit3}
\end{figure}

\paragraph{Stronger DeiT-III training.}
We additionally repeat the evaluation using the stronger DeiT-III training
recipe~\citep{touvron2022deit}. \emph{Procedural Core} does not improve
ImageNet classification ($82.4\%$ versus $82.5\%$ top-1), is approximately
level on \textsc{ADE20K}, and is slightly worse on \textsc{NYUv2}.
However, improvements persist on the two attention-based tasks, with
$+2.07$ mAP on \textsc{ImageNet-S} and $+7.70$ corloc on \textsc{VOC07}
(Figure~\ref{fig:deit3}).

\section{Extended Discussion and Limitations}
\label{app:extended_discussion}

\paragraph{Why study standard architectures?}
Our goal in this work is not to develop a training recipe specialized to a particular state-of-the-art architecture, but to understand a more fundamental question: whether useful computational structure can be introduced directly through a transformer's initialization, and how that structure influences subsequent learning. We therefore focus primarily on the standard ViT architecture, whose simplicity and extensive study provide a controlled setting for isolating the effect of initialization and performing targeted mechanistic interventions.

This setting also allows us to examine the same initialization across a broad range of applications. Beyond image classification, we study self-supervised learning, semantic and zero-shot segmentation, object localization, and depth estimation, as well as language modeling. Our analysis identifies recurrence and structure in the attention value/output pathway as important contributors to transfer. We expect these principles to help guide analogous approaches for more modern architectures such as XCiT and other transformer variants. Determining how the learned structure should be adapted to their architectural components is left to future work.

\paragraph{Interaction with modern training recipes.}
We additionally evaluate \emph{Procedural Core} under the substantially stronger DeiT-III training recipe. In this setting, it does not improve ImageNet-1K classification, while gains persist on ImageNet-S zero-shot segmentation and VOC07 unsupervised object localization. DeiT-III simultaneously changes several aspects of downstream training, including optimization, augmentation, and regularization, and is itself a heavily optimized recipe for image classification.

Importantly, we do not retune this recipe specifically for models initialized with \emph{Procedural Core}. The optimal training protocol for a structured initialization need not coincide with one developed for random initialization, and DeiT-III therefore should not be viewed as an upper bound on the potential benefit of the method. Alternative optimization, augmentation, or regularization choices may better preserve or amplify the computational structure introduced at initialization. Understanding these interactions, and potentially jointly designing initialization and downstream training recipes, is an important direction for future work.

\end{document}